\documentclass[9pt,journal]{IEEEtran}
\IEEEoverridecommandlockouts

\usepackage{color}

\usepackage{placeins}
\usepackage{amsthm}
\usepackage{mathtools}
\usepackage{amssymb,amsmath}
\usepackage[linesnumberedhidden,ruled,vlined]{algorithm2e}
\usepackage{algpseudocode}

\usepackage{graphicx}

\usepackage{textcomp}

\usepackage{multirow}
\usepackage[table]{xcolor}
\usepackage{threeparttable}

\usepackage{array}
\newcolumntype{?}{!{\vrule width 1pt}}

\usepackage{verbatim}
\usepackage{cite}
\usepackage{url}
\usepackage{cases}

\usepackage{capt-of}

\usepackage{upgreek}
\usepackage{bm}
\usepackage{nicefrac}
\usepackage[normalem]{ulem}

\ifCLASSOPTIONcompsoc
  \usepackage[caption=false,font=normalsize,labelfont=sf,textfont=sf]{subfig}
\else
  \usepackage[caption=false,font=footnotesize]{subfig}
\fi

\usepackage{cite}
\usepackage{balance}

\renewcommand{\IEEEQED}{\IEEEQEDopen}
\let\oldnl\nl
\newcommand{\nonl}{\renewcommand{\nl}{\let\nl\oldnl}}

\newdimen\origiwspc%
\newdimen\origiwstr%
\origiwspc=\fontdimen2\font
\origiwstr=\fontdimen3\font

\IEEEoverridecommandlockouts

\begin{document}
\title{Dictionary Learning for Adaptive GPR Landmine Classification}

\author{\fontdimen2\font=0.5ex{Fabio~Giovanneschi,
		Kumar Vijay~Mishra, 		
        Maria Antonia~Gonzalez-Huici,
        Yonina C.~Eldar
        and Joachim H. G.~Ender}\fontdimen2\font=\origiwspc
\thanks{F. G. and M. A. G.-H. are with the Fraunhofer Institute for High Frequency Physics and Radar Techniques, Bonn, Germany. E-mail: \{fabio.giovanneschi, maria.gonzalez\}@fhr.fraunhofer.de.}
\thanks{K. V. M. is with The University of Iowa, Iowa City, Iowa, USA. E-mail: kumarvijay-mishra@uiowa.edu.}
\thanks{Y. C. E. is with the Weizmann Institute of Science, Rehovot, Israel. E-mail: yonina.eldar@weizmann.ac.il.}
\thanks{J. H. G. E. is with the Centre for Sensor Systems (ZESS), University of Siegen, Germany. E-mail: ender@zess.uni-siegen.de.}
\thanks{The authors would like to thank the colleagues at Leibniz Institute for Applied Geophysics (LIAG), Hannover, Germany for their support during the measurement campaign. K.V.M. acknowledges partial support via Lady Davis Postdoctoral Fellowship and Andrew and Erna Finci Viterbi Postdoctoral Fellowship.}}

\maketitle
\begin{abstract}
Ground penetrating radar (GPR) target detection and classification is a challenging task. Here, we consider online dictionary learning (DL) methods to obtain sparse representations (SR) of the GPR data to enhance feature extraction for target classification via support vector machines. Online methods are preferred because traditional batch DL like K-SVD is not scalable to high-dimensional training sets and infeasible for real-time operation. We also develop Drop-Off MINi-batch Online Dictionary Learning (DOMINODL) which exploits the fact that a lot of the training data may be correlated. The DOMINODL algorithm iteratively considers elements of the training set in small batches and drops off samples which become less relevant. For the case of abandoned anti-personnel landmines classification, we compare the performance of K-SVD with three online algorithms: classical Online Dictionary Learning, its correlation-based variant, and DOMINODL. Our experiments with real data from L-band GPR show that online DL methods reduce learning time by 36-93\% and increase mine detection by 4-28\% over K-SVD. Our DOMINODL is the fastest and retains similar classification performance as the other two online DL approaches. We use a Kolmogorov-Smirnoff test distance and the Dvoretzky-Kiefer-Wolfowitz inequality for the selection of DL input parameters leading to enhanced classification results. 
To further compare with state-of-the-art classification approaches, we evaluate a convolutional neural network (CNN) classifier which performs worse than the proposed approach. Moreover, when the acquired samples are randomly reduced by 25\%, 50\% and 75\%, sparse decomposition based classification with DL remains robust while the CNN accuracy is drastically compromised.
\end{abstract}


\IEEEpeerreviewmaketitle

\section{Introduction}
\label{sec:intro}
A ground penetrating radar (GPR, hereafter) is used for probing the underground by transmitting radio waves from an antenna held closely to the surface and acquiring the echoes reflected from subsurface anomalies or buried objects. As the electromagnetic wave travels through the subsurface, its velocity changes due to the physical properties of the materials in the medium. By recording such changes in the velocity and measuring the travel time of the radar signals, a GPR generates profiles of scattering responses from the subsurface. The interest in GPR is due to its ability to reveal buried objects non-invasively and detect non-metallic scatterers with increased sensitivity to dielectric contrast \cite{jol2008ground}. 
From the recordings of the previously observed regions, GPR surveys can also extrapolate subsurface knowledge for inaccessible or unexcavated areas. This sensing technique is, therefore, attractive for several applications such as geophysics, archeology, forensics, and defense (see e.g. \cite{daniels2005ground,jol2008ground} for some surveys). Over the last decade, there has been a spurt in GPR research because of advances in electronics and computing resources. GPR has now surpassed traditional ground applications and has become a more general ultra-wideband remote sensing system with proliferation to novel avenues such as through-the-wall imaging, building construction, food safety monitoring, and vegetation observation. 

In this work, we consider the application of detecting buried landmines using GPR. This is one of the most extensively investigated GPR applications due to its obvious security and humanitarian importance\cite{monitor2013landmine}. Mine detection GPR usually operates in the L-band ($1$-$2$ GHz) with ultra-wideband (UWB) transmission in order to achieve both sufficient resolution to detect small targets ($5$-$10$ cm diameter) and penetrate solid media at shallow depths ($15$-$30$ cm) \cite{giovanneschi2013parametric}.

Even though a lot of progress has been made on GPR for landmine detection, discriminating them from natural and manmade clutter remains a critical challenge. In such applications, the signal distortion due to inhomogeneous soil clutter, surface roughness and antenna ringing hampers target recognition. Moreover, the constituting material of many models of landmines is largely plastic and has a very weak response to radar signals due to its low dielectric contrast with respect to the soil \cite{daniels2005ground}. 
Finally, a major problem arises due to low radar cross section (RCS) of some landmine models \cite{gonzalez2014comparative}. A variety of signal processing algorithms have been proposed for detection of low metal-content landmines in realistic scenarios; approaches based on feature extraction and classification are found to be the most successful (see e.g. \cite{gonzalez2013combined,torrione2014histograms,giannakis2016model}), yet false-alarm rates remain very high. 


Sparse representation (SR)
is effective in extracting the mid- or high-level features in image classification \cite{wright2009robust,wright2010sparse}. In the context of anti-personnel landmines recognition using GPR, our prior work \cite{giovanneschi2015preliminary,giovanneschi2017online} has shown that frameworks based on SR improve the performance of Support Vector Machine (SVM) classifiers in distinguishing different types of mines and clutter in highly corrupted GPR signals.
In this approach, the signal-of-interest is transformed into a domain where it can be expressed as a linear combination of only a few \textit{atoms} chosen from a collection called the \textit{dictionary} matrix \cite{elad2006image,eldar2015sampling}. The dictionary may be learned from the data it is going to represent. \textit{Dictionary learning} (DL) techniques (or \textit{sparse coding} in machine learning parlance) aim to create adaptive dictionaries which provide the sparsest reconstruction for given training-sets, i.e., a representation with a minimum number of constituting atoms. DL methods are critical building blocks in many applications such as deep learning, image denoising, and super-resolution; see \cite{elad2010prologue,arora2014new,eldar2012compressed} for further applications. 

Classical DL algorithms such as Method of Optimal Directions (MOD) \cite{engan1999method} and K-SVD \cite{elad2006image} operate in batches - dealing with the entire training set in each iteration. Although extremely successful, these methods are computationally demanding and not scalable to high-dimensional training sets. An efficient alternative is the Online Dictionary Learning (ODL) algorithm \cite{mairal2009online} that has faster convergence than batch DL methods. 
In this work, we develop a new approach toward classification based on online DL-SR framework for the specific case of GPR-based landmine identification.

The main contributions of this paper are as follows:\\
\textbf{1) Faster DL for GPR landmine classification.}\footnote{The conference precursor of this work was presented in IEEE International Geoscience and Remote Sensing Symposium, 2017 \cite{giovanneschi2017online}.} 
We investigate the application of DL towards GPR-based landmine classification. To the best of our knowledge, this has not been investigated previously. Furthermore, online DL\footnote{We use the term ``online DL'' to imply any algorithm that operates in online mode. From here on, we reserve the term ODL solely to refer to the method described in \cite{mairal2009online}.} methods have been studied more generally in GPR. Only one other previous study has employed DL (K-SVD) using GPR signals \cite{shao2013sparse}, although for the application of identifying bedrock features. We employ online DL methods and use the coefficients of the resulting sparse vectors as input to a SVM classifier to distinguish mines from clutter. Our comparison of K-SVD and online DL using real data from L-band GPR shows that online DL algorithms present distinct advantages in speed and low false-alarm rates.
We propose a new Drop-Off MINi-batch Online Dictionary Learning (DOMINODL) which processes the training data in mini-batches and avoids unnecessary update of the irrelevant atoms in order to reduce the computational complexity. The intuition for the dropoff step comes from the fact that some training samples are highly correlated and, therefore, in the interest of processing time, they can be dropped during training without significantly affecting performance.\\
\textbf{2) Better statistical metrics for improved classification.} Contrary to previous studies \cite{shao2013sparse} which determine DL parameters (number of iterations, atoms, etc.) based on bulk statistics such as normalized root-mean-square-error (NRMSE), we consider statistical inference for parameter analysis. Our methods based on Kolmogorov-Smirnoff test distance \cite{chakravarti1967handbook} and Dvoretzky-Kiefer-Wolfowitz (DKW) Inequality \cite{dvoretzky1956asymptotic,massart1990tight} are able to fine-tune model selection resulting in improved mine classification performance.\\
\textbf{3) Experimental validation for different landmine sizes.} Our comparison of K-SVD with three online DL algorithms - ODL, its correlation-based variant \cite{naderahmadian2016correlation} and DOMINODL - shows that online methods successfully detect mines with very small RCS buried deep into clutter and noise. Some recent studies \cite{wilson2007large,robledo2009survey,lameri2017landmine,besaw2015deep} employ state-of-the-art deep learning approaches such as a convolutional neural network (CNN) to classify GPR-based mines data. Our comparison with CNN illustrates that it has poorer performance in detection of small mines than our online DL approaches. This may also be caused by the relatively small dimensions of our training set which, even if perfectly adequate for DL, may not meet the expected requirements for CNN \cite{sontag1998vc}. We also show that the classification performance of online DL methods does not deteriorate significantly when signal samples are reduced.

The rest of the paper is organized as follows. In the next section, we formally describe the classification problem and GPR specific challenges. In Section~\ref{sec:prop_meth}, we explain various DL algorithms used in our methodology and also describe DOMINODL. We provide an overview of the GPR system and field campaign to collect GPR data sets in Section~\ref{sec:meas_camp}. In Section~\ref{sec:param}, we introduce our techniques for DL parameter selection. Section~\ref{sec:exp} presents classification and reconstruction results using real radar data. We conclude in Section~\ref{sec:summary}.

Throughout this paper, we reserve boldface lowercase and uppercase letters for vectors and matrices, respectively. The $i$th element of a vector $\textbf{y}$ is $\mathbf{y}_i$ while the $(i,j)$th entry of the matrix $\textbf{Y}$ is $\textbf{Y}_{i,j}$. We denote the transpose by $(\cdot)^T$. We represent the set of real and complex numbers by $\mathbb{R}$ and $\mathbb{C}$, respectively. Other sets are represented by calligraphic letters. The notation $\left\Vert\cdot\right\Vert_p$ stands for the $p$-norm of its argument and $\left\Vert\cdot\right\Vert_F$ is the Frobenius norm. A subscript in the parenthesis such as $(\cdot)_{(t)}$ is the value of the argument in the $t$th iteration. The  convolution product is denoted by $\ast$. The function $\text{diag}(\cdot)$ outputs a diagonal matrix with the input vector along its main diagonal. We use $\text{Pr}\{\cdot\}$ to denote probability, $\text{E}\left\lbrace \cdot \right\rbrace $ is the statistical expectation, and $|\cdot|$ denotes the absolute value. The functions $\text{max}(\cdot)$ and $\text{sup}(\cdot)$ output the maximum and supremum value of their arguments, respectively.

\section{Problem Formulation}
\label{sec:prob_form}
A pulsed GPR transmits a signal into the ground and receives its echo for each point in the radar coverage area. The digital samples of this echo constitute a \textit{range profile} designated by the signal vector $\mathbf{y} \in \mathbb{R}^M$ where $M$ is the number of range cells. Formally, the SR of $\mathbf{y}$ can be described by $\mathbf{y} = \mathbf{D}\mathbf{x}$, where $\mathbf{D} = [\mathbf{d}_1,\cdots,\mathbf{d}_N] \in \mathbb{R}^{M\times N}$ with column vectors or \textit{atoms} $\{\mathbf{d}_i\}$ is a redundant or overcomplete ($M \ll N$) dictionary, and $\mathbf{x} \in \mathbf{R}^N$ is the sparse representation vector. The SR process finds the decomposition that uses the minimum number of atoms to express the signal. If there are $L$ range profiles available, then the SR of the data $\mathbf{Y} = [\mathbf{y}_1, \cdots, \mathbf{y}_L]$ is described as $\mathbf{Y} = \mathbf{D}\mathbf{X}$, where $\mathbf{X} = [\mathbf{x}_1, \cdots, \mathbf{x}_L] \in \mathbb{R}^{N\times L}$.

Our goal is to classify different mines (including the ones with small RCS) and clutter based on the SR of range profiles using fast, online DL methods. The GPR range profiles from successive scans are highly correlated. We intend to exploit this property during the DL process. In the following, we describe the SR-based classification and mention its challenges.

\subsection{GPR Target Classification Method}
\label{subsec:GPRCLASS}
We use SVM to classify sparsely represented GPR range profiles using the learned dictionary {$\mathbf{D}$}. Given a predefined collection of labeled observations, SVM searches for a functional $f:\mathbb{R}^n \rightarrow \mathbb{R}$ that maps any new observation to a class $c\in \mathbb{R}$. A binary classifier with linearly separable data, for example, would have $c \in \{1, -1\}$. In our work, we use {$\mathbf{X}$}, i.e. the sparse decomposition of a given set of signals (the `training' signals) {$\mathbf{Y}$} using the learned dictionary {$\mathbf{D}$}, as a set of labeled observations for the SVM. SVM transforms the data into a high dimensional feature space where it is easier to separate between different classes. The kernel function that we use to compute the high dimensional operations in the feature space, is the Gaussian Radial Basis Function (RBF): $\kappa(\mathbf{x}_i,\mathbf{x}_j) = \exp(-\gamma(||\mathbf{x}_i-\mathbf{x}_j||)^2 + C)$, where $\gamma > 0$ is the free parameter that decides the influence of the support data vector $\mathbf{x}_j$ on the class of the vector $\mathbf{x}_i$ in the original space and $C$ is the parameter for the soft margin cost function which controls the influence of each individual support vector \cite{chang2011libsvm}. To optimally select the SVM input parameters, we arrange the original classification set into training and validation vectors in $\nu$ different ways ($\nu$-fold cross-validation with $\nu$=10) to arrive at a certain mean cross-classification accuracy of the validation vectors. The folds were randomly selected and their number was found empirically by determining the limit above which the accuracy improvement was negligible. We refer the reader to \cite{chang2011libsvm} for more details on SVM.

\subsection{The Basic Framework of DL}
\label{subsec:dlbasics}
In many applications, the dictionary $\mathbf{D}$ is unknown and has to be learned from the training signals coming from the desired class. A DL algorithm finds an over-complete dictionary $\mathbf{D}$ that sparsely represents measurements $\mathbf{Y}$ 
such that $\mathbf{Y} \simeq \mathbf{D}\mathbf{X}$. Each of the vectors $\mathbf{x}_i$ is a sparse representation of $\mathbf{y}_i$ with only $K_s$ nonzero entries. A non-tractable formulation of this problem is
\begin{flalign}
\label{eq:csrecoverl1_1}
	& \underset{\mathbf{D},\mathbf{X}}{\text{minimize}}\phantom{1}\left\Vert \mathbf{Y}-\mathbf{D}\mathbf{X}\right\Vert _{F}\nonumber\\
	& \text{subject to}\phantom{1} \left\Vert\mathbf{x}_i\right\Vert_0 \le K_s,\: 1\le i \le L.
\end{flalign}
Since both $\mathbf{D}$ and $\mathbf{X}$ are unknown, a common approach is to use alternating minimization in which we start with an initial guess of $\mathbf{D}$ and then obtain the solution iteratively by alternating between two stages - \textit{sparse coding} \cite{aharon2006k} and \textit{dictionary update} \cite{lewicki2000learning} - as follows: \\
\textit{1) Sparse Coding}: Obtain $\mathbf{X}_{(t)}$ as:
\begin{align}
\label{eq:spcoding1}
	\mathbf{X}_{(t)} &= \underset{\mathbf{X}}{\text{minimize}}\phantom{1}\left\Vert \mathbf{Y}-\mathbf{D}_{(t-1)}\mathbf{X}\right\Vert _{F}\nonumber\\
	& \text{subject to}\phantom{1} \left\Vert\mathbf{x}_{i_{t-1}}\right\Vert_p \le K_s,\: 1\le i \le L,
\end{align}
where $\mathbf{X}_{(t)}$ is the SR in $t^{th}$ iteration. This can be solved using greedy algorithms such as orthogonal matching pursuit (OMP) ($p=0$), convex relaxation methods like basis pursuit denoising (BPDN) ($p=1$) or focal underdetermined system solver (FOCUSS) ($0<p<1$).\\
\textit{2) Dictionary Update}: Given $\mathbf{X}_{(t)}$, update $\mathbf{D}$ such that
\begin{align}
\label{eq:dl update1}
	\mathbf{D}_{(t)} &= \underset{\mathbf{D}\in\mathcal{D}}{\text{minimize}}\phantom{1}\left\Vert \mathbf{Y}-\mathbf{D}\mathbf{X}_{(t)}\right\Vert _{F},
\end{align}
where $\mathcal{D}$ is a set of all dictionaries with unit column-norms, $\left\Vert\mathbf{d}_j\right\Vert_2 = 1$ for $1\le j\le N$. This subproblem is solved by methods such as singular value decomposition or gradient descent \cite{aharon2006k,mairal2009online}.

Classical methods such as MOD \cite{engan1999method} and K-SVD \cite{elad2006image} retain a guess for $\mathbf{D}$ and $\mathbf{X}$ and iteratively update either $\mathbf{X}$ using basis/matched pursuit (BP/MP) or $\mathbf{D}$ using least squares solver. Both MOD and K-SVD operate in batches, i.e. they deal with the entire training set in each iteration, and solve the same dictionary learning model but differ in the optimization method. Since the initial guesses of $\mathbf{D}$ or $\mathbf{X}$ can be far removed from the actual dictionary, the BP step may behave erratically. 
While there are several state-of-the-art results that outline DL algorithms with concrete performance guarantees, e.g. \cite{spielman2012exact,agarwal2014learning,arora2014new}, they require stronger assumptions on the observed data. In practice, heuristic DL such as MOD and K-SVD do yield overcomplete dictionaries although provable guarantees for such algorithms are difficult to come by. 

Several extensions of batch DL have been proposed e.g. label consistent (LC) K-SVD \cite{jiang2013label} and discriminative K-SVD \cite{zhang2010discriminative} introduced label information into the procedure of learning dictionaries to make them more discriminative. The performance of K-SVD can be improved in terms of both computational complexity and obtaining an incoherent dictionary if the learning process enforces constraints such as hierarchical tree sparsity \cite{varshney2008sparse}, structured group sparsity (StructDL) \cite{suo2014structured}, Fisher discrimination (FDDL) \cite{yang2014sparse}, and low-rank-and-Fisher (D$^2$L$^2$R$^2$) \cite{li2014learning}. 
Often objects belonging to different classes have common features. This has been exploited in improving K-SVD to yield methods such as DL with structured incoherence and shared features (DLSI) \cite{ramirez2010classification}, separating the commonality and the particularity (COPAR) \cite{kong2012dictionary}, convolutional sparse DL (CSDL) \cite{gao2014learning}, shift-invariant DL \cite{rusu2018learning}, principal component analysis DL \cite{nguyen2018separation}, convolutional DL \cite{garcia2018convolutional} and low-rank shared DL (LRSDL) \cite{vu2017fast}. A recent review of various DL algorithms can be found in \cite{dumitrescu2018dictionary}.

In general, batch DL methods are computationally demanding at test time and not scalable to high-dimensional training sets. On the other hand, online methods such as ODL \cite{mairal2009online} converge fast and process small sets. A few improvements to ODL have already been proposed. For example, \cite{sulam2016trainlets} considered a faster Online Sparse Dictionary Learning (OSDL) to efficiently handle bigger training set dimensions using a double-sparsity model. A recent study \cite{naderahmadian2016correlation} notes that even though online processing reduces computational complexity compared to batch methods, ODL performance can be further improved if the useful information from previous data is not ignored in updating the atoms. In this study, a new online DL called Correlation-Based Weighted Least Square Update (CBWLSU) was proposed, which employs only part of the previous data correlated with current data for the update step. The CBWLSU is relevant to GPR because the latter often contains highly correlated range profiles.

In this paper, our focus is to investigate such fast DL methods in the context of GPR-based landmine detection and classification. We also propose a new online DL method that exploits range profile correlation as in CBWLSU but is faster than both ODL and CBWLSU. Our inspiration is the K-SVD variant called incremental structured DL (ISDL) that was used earlier in the context of SAR imaging \cite{chen2016unsupervised}. In ISDL, at each iteration, a small batch of samples is randomly drawn from the training set. Let $\mathcal{R}_t \subset \{1,\cdots,L\} $ be the set of indices of the mini-batch training elements chosen uniformly at random at the $t^{\text{th}}$ iteration. Then, ISDL updates the dictionary $\mathbf{D}$ using the mini-batch $\mathbf{Y}_{\mathcal{R}_t} = \{\mathbf{y}_l: l \in \mathcal{R}_t\}$ and the corresponding representation coefficient $\mathbf{X}$. The fast iterative shrinkage-thresholding algorithm (FISTA) \cite{beck2009fast} and block coordinate descent methods solve the sparse coding and dictionary update, respectively. As we will see in later sections, the mini-batch strategy that we employed in our DOMINODL reduces computational time without degrading performance. 

\section{Online DL}
\label{sec:prop_meth}
We now describe the DL techniques used for GPR target classification and then develop DOMINODL in order to address challenges of long training times in the context of our problem. 

\subsection{K-SVD, LRSDL, ODL and CBWLSU}
\label{subsec:ksvd_odl}
As mentioned earlier, the popular K-SVD algorithm \cite{rubinstein2008efficient} sequentially updates all the atoms during the dictionary update step using all training set elements. For the sparse coding step at iteration $t$, K-SVD employs OMP with the formulation: 
\begin{flalign}
\label{eq:omp}
& \underset{\mathbf{x}_i}{\text{minimize}}\phantom{1}\left\Vert\mathbf{x}_i\right\Vert_0\nonumber\\
	& \text{subject to}\phantom{1} \left\Vert \mathbf{y}_i-\mathbf{D_{(t-1)}}\mathbf{x}_i\right\Vert^2_{2} \le \delta,\:\forall 1\le i \le L,
\end{flalign}
where $\delta$ is the maximum residual error used as a stopping criterion. For the dictionary update at iteration $t$, K-SVD solves the global minimization problem (\ref{eq:dl update1}) via $K$ sequential minimization
problems, wherein every column $\mathbf{d}_k$ of $\mathbf{D}$ and its corresponding row of coefficients $\mathbf{X}_{\text{row},k}$ of $\mathbf{X}$ are updated, as follows
\begin{flalign}
\label{eq:ksvd_du}
	&\{\mathbf{X}_{\text{row},k_{(t)}}, \mathbf{d}_{k_{(t)}}\} =\nonumber\\
    &\underset{\mathbf{X}_{\text{row},k_{(t-1)}}, \mathbf{d}_{k}}{\text{minimize}}\phantom{1}\left\Vert \mathbf{Y}- \sum\limits_{l\neq k}\mathbf{d}_{l_{t-1}}\mathbf{X}_{\text{row},l_{(t-1)}} -\mathbf{d}_{k}\mathbf{X}_{\text{row},k_{(t-1)}}\right\Vert _{F}.
\end{flalign}
The update process employs SVD to find the closest rank-1 approximation (in Frobenius norm) of the error term $\mathbf{Y}- \sum\limits_{l\neq k}\mathbf{d}_{l_{t-1}}\mathbf{X}_{\text{row},l_{(t-1)}}$ subject to the constraint $\Vert\mathbf{d}_{k_{(t)}}\Vert_2 = 1$.\\ 

Another recent batch method of interest is the low-rank shared DL (LRSDL) \cite{vu2017fast}. This is a discriminative batch DL algorithm (others being D-KSVD \cite{zhang2010discriminative} and LC-KSVD \cite{jiang2013label}) that learns by promoting the generation of a dictionary $\mathbf{D}$ which is separated in blocks of atoms associated to different classes as $\mathbf{D} = [\mathbf{D}_1, \cdots, \mathbf{D}_C] \in \mathbb{R}^{M\times N}$ where $C$ is the number of classes present in the training set $\mathbf{Y}$. The resultant coefficient matrix $\mathbf{X}$ has a block diagonal structure. The assumption of non-overlapping subspaces is often unrealistic in practice. Techniques such as COPAR \cite{kong2012dictionary}, JDL \cite{zhou2014jointly} and CSDL \cite{gao2014learning} exploit common patterns among different classes even though different objects possess distinct class-specific features. These methods produce am additional constituent $\mathbf{D_0}$ which is shared among all classes so that $\mathbf{D} = [\mathbf{D}_1, \cdots, \mathbf{D}_C, \mathbf{D}_0] \in \mathbb{R}^{M\times N}$. The drawback of these strategies is that the shared dictionary may also contain class-discriminative features. To avoid this problem, LRSDL requires that the shared dictionary must have a low-rank structure and that its sparse coefficients have to be almost similar. Once the data is sparsely represented with such dictionaries, a sparse-representation-based classifier (SRC) is used to predict the class of new data. The LRSDL update process employs alternating direction method of multipliers (ADMM) \cite{boyd2011distributed} and FISTA for the sparse decomposition step.

ODL is an interesting alternative for inferring a dictionary from large training sets or ones which change over time \cite{mairal2009online}. ODL also updates the entire dictionary sequentially, but uses one element of training data at a time for the dictionary update. Assuming that the training set $\mathbf{Y}$ is composed of independent and identically distributed samples of a distribution $p(\mathbf{Y})$, ODL first draws an example of the training set $\mathbf{y}_{t}$ from $\mathbf{Y}$. Then, the sparse coding is the Cholesky-based implementation of the LARS-LASSO algorithm \cite{osborne2000new}. The latter solves a $\ell_1$-regularized least-squares problem for each column of $\mathbf{Y}$. In the dictionary update we consider all the training set elements analyzed so far, namely, $\mathbf{y}_i \phantom{1} \text{with} \phantom{1} i = 1 ... t$:
\begin{align}
\label{lars1}
\mathbf{x}_{t} = \underset{\mathbf{x} \in \mathbb{C}^n }{\mathrm{minimize}}\phantom{1} \frac{1}{2}||\mathbf{y}_t - \mathbf{D_{(t-1)}}\mathbf{x}||^2_2 + \lambda||\mathbf{x}||_1.
\end{align}
In the next step, each column of $\mathbf{D}$ is sequentially updated via gradient descent using the dictionary computed in the previous iteration. Before receiving the next training data, the dictionary update is repeated multiple times for convergence.

CBWLSU is an online method that introduces an interesting alternative for the dictionary update step \cite{naderahmadian2016correlation}. Like ODL, CBWLSU evaluates one new training data $\mathbf{y}_i$. However, to update the dictionary, it searches among all previous training data and uses only the ones which share the same atoms with $\mathbf{y}_{i}$. Let $\mathbf{Y}_{\mathcal{Q}_t} =  \{\mathbf{y}_l: l \in \mathcal{Q}_t\}$ with $\mathcal{Q}_t = \{\mathbf{y}_l: 1 \le l < t-1\}$ be the set of previous training elements at iteration $i$. Define $\mathcal{N}_t = \left\lbrace l : 1 < l < t, \left\langle \mathbf{x}_l^T \cdot \mathbf{x}_t \right\rangle \neq 0 \right\rbrace \subset \mathcal{Q}_t$ as the set of indices of all previous training elements that are correlated with the new element such that $|\mathcal{N}_t| = N_{p_t}$. The new training set is $\mathbf{Y}_{\mathcal{N}_t} = \{\mathbf{y}_l: l \in \mathcal{N}_t\} \cup \mathbf{y}_t$. Then, CBWLSU employs a weighting matrix $\mathbf{W}(\mathbf{y}_t)$ to evaluate the influence of the selected previous elements $\mathbf{y}_t$ for the dictionary update step and solves the optimization problem therein via weighted least squares. Unlike K-SVD and ODL, CBWLSU does not require the dictionary pruning step to replace the unused or rarely used atoms with the training data. The sparse coding in CBWLSU is achieved via batch OMP.

\subsection{Drop-Off Mini-Batch Online Dictionary Learning}
\label{subsec:dl_gprs}

\begin{table*}
\centering
 	\caption{Comparison of DL steps} 
	\label{tbl:dlcomparison}
	\begin{tabular}{ l l l l l l }
		\hline
         \noalign{\vskip 1pt}    
         	 DL step & K-SVD &  LRSDL & ODL & CBWLSU   & DOMINODL\\[1pt]
		\hline
		\hline
        \noalign{\vskip 1pt}    
		   Training method & Batch & Batch & Online & Online  & Online\\[1pt]
            Sparse coding method & OMP & FISTA & LARS & Batch OMP & Entropy-thresholded batch OMP\\[1pt]  
            Dictionary update & Entire $\mathbf{D}$ atom-wise & Entire $\mathbf{D}$ & Entire $\mathbf{D}$ group-wise & Entire $\mathbf{D}$ atom-wise   & Partial $\mathbf{D}$ adaptively group-wise\\[1pt]
            Training samples per iteration & Entire $\mathbf{Y}$ & Entire $\mathbf{Y}$ & $\mathbf{Y}_{t}$ & $\mathbf{Y}_{\mathcal{N}_t}$  & $\mathbf{Y}_{\mathcal{C}_t}$\\[1pt]
            Optimization method &  SVD & ADMM &  Gradient descent & Weighted least squares  & Weighted least squares\\[1pt]
            Post-update dictionary pruning &  Yes & No &  Yes & No & No\\[1pt]
             Training-set drop-off &  No &  No &  No & No & Yes\\[1pt]
		\hline
		\hline
        \noalign{\vskip 2pt}
	\end{tabular}
\end{table*}  

We now introduce our DOMINODL approach for online DL which not only leads to a dictionary ($\mathbf{D}$) that is tuned to sparsely represent the training set ($\mathbf{Y}$) but is also faster than other online algorithms. The key idea of DOMINODL is as follows: When sequentially analyzing the training set, it is pertinent to leverage the memory of previous data in the dictionary update step. However, algorithms such as CBWLSU consider \textit{all} previous elements. Using all previous training set samples is computationally expensive and may also slow down convergence. The samples which have already contributed in the dictionary update do not need to be considered again. Moreover, in some real-time applications (such as highly correlated range profiles of GPR), their contribution may not be relevant anymore for updating the dictionary. 

In DOMINODL, we save computations by considering only a small batch of previous elements that are correlated with the new elements. The two sets are defined correlated if, in their sparse decomposition, they have at least one common non-zero element. The time gained from considering fewer previous training elements is used to consider a mini-batch of new training data (instead of a single element as in ODL and CBWLSU). The sparse coding step of DOMINODL employs batch OMP, selecting the maximal residual error $\delta$ in (\ref{eq:omp}) using a data-driven entropy-based strategy as described later in this section. At the end of each iteration, DOMINODL also drops-off those previous training set elements that have not been picked up after a certain number of iterations, $N_u$. The mini-batch drawing combined with dropping off training elements and entropy-based criterion to control sparsity results in an extremely fast online DL algorithm that is beneficial for real-time radar operations.

We initialize the dictionary $\mathbf{D}$ using a collection of $K$ training set samples that are randomly chosen from $\mathbf{Y}$. We then perform a sparse decomposition of $\mathbf{Y}$ with the dictionary $\mathbf{D}$. 
Let the iteration count $t$ indicate the $t^{\text{th}}$ element of the training set. We define the mini-batch of $N_b$ new training elements as $\mathbf{Y}_{\mathcal{B}_t} =  \{\mathbf{y}_l: l \in \mathcal{B}_t\}$, where $\mathcal{B}_t = \{\mathbf{y}_l: i \le l < t+N_b\}$ with $\mathcal{B}_t \subset \mathcal{N}_t$ and $|\mathcal{B}_t| = N_b$. When $t>L-N_b$, we simply take the remaining new elements to constitute this mini-batch\footnote{In numerical experiments, we observed that the condition $t>L-N_b$ rarely occurs because DOMINODL updates the dictionary and converges in very few iterations. The algorithm also ensures that the number of previous samples $\geq 2N_r$ before the dictionary update. If this condition is not fulfilled, then it considers all previous training samples.}. We store the set of dictionary atoms participating in the SR of the signals in $\mathbf{Y}_{\mathcal{B}_t}$ as $\mathbf{D}_{\mathcal{B}_t}$. We define $\mathbf{Y}_{\mathcal{Q}_t} =  \{\mathbf{y}_l: l \in \mathcal{Q}_t\}$ with $\mathcal{Q}_t = \{\mathbf{y}_l: 1 \le l < t-1\}$ as the set of previous training elements at iteration $t$. We consider a randomly selected mini-batch $\mathbf{Y}_{\mathcal{M}_t}= \{\mathbf{y}_l: l \in \mathcal{M}_t\}$ with $\mathcal{M}_t \subset \mathcal{Q}_t$ such that $|\mathcal{M}_t| = N_r$. Let $\mathbf{Y}_{\mathcal{A}_t} = \{\mathbf{y}_l: l \in \mathcal{A}_t\}$ where $\mathcal{A}_t \subset \mathcal{M}_t$ such that $\mathcal{A}_t = \left\lbrace l : l \in \mathcal{M}_t, \left\langle \mathbf{x}_l^T \cdot \mathbf{x}_t \right\rangle \neq 0 \right\rbrace$ be a subset of previous training elements that are correlated with the mini-batch of new elements. In order to avoid multiple occurrences of the same element in consecutive mini-batches, DOMINODL ensures that $\mathcal{M}_t \cap \mathcal{M}_{t-1} = \emptyset$. Let $\mathbf{D}_{\mathcal{A}_t}$ be the set of dictionary atoms used for SR of $\mathbf{Y}_{\mathcal{A}_t}$. Our new training set is $\mathbf{Y}_{\mathcal{C}_t} = \mathbf{Y}_{\mathcal{A}_t} \cup \mathbf{Y}_{\mathcal{B}_t}$. Both mini-batches of new and previous elements are selected such that the entire training set size ($N_b+N_r$) is still smaller than that of CBWLSU where it is $N_{p_t}+1$.\\

The dictionary update subproblem reduces to considering only the sets $\mathbf{Y}_{\mathcal{C}_t}$, $\mathbf{D}_{\mathcal{C}_t}$ and $\mathbf{X}_{\mathcal{C}_t}$:
\begin{align}
\label{eq:DL_DOMINODL}
\mathbf{\hat{D}}_{\mathcal{C}_t} = \underset{\mathbf{D}_{\mathcal{C}_t} \in \mathcal{D}}{\text{minimize}} \phantom{1} || \mathbf{Y}_{\mathcal{C}_t}-\mathbf{D}_{\mathcal{C}_t}\mathbf{X}_{\mathcal{C}_t}||_{F}^2.
\end{align}
\normalsize
Assume that the sparse coding for each example is known and define the errors as
\begin{align}
\label{eq:Error_DOMINODL}
\mathbf{E}_{\mathcal{C}_t} = \mathbf{Y}_{\mathcal{C}_t} -\mathbf{D}_{\mathcal{C}_t}\mathbf{X}_{\mathcal{C}_t} = [\mathbf{e}_1, \cdots,\mathbf{e}_{N_r}].
\end{align}
\normalsize
We can update $\mathbf{D}_{\mathcal{C}_t}$, such that the above error is minimized, with the assumption of fixed $\mathbf{X}_{\mathcal{C}_t}$. A similar problem is considered in MOD where error minimization is achieved through least squares. Here, we employ weighted least squares inspired by the fact that it has shown improvement in convergence over standard least squares \cite{naderahmadian2016correlation}. We compute the weighting matrix $\mathbf{W}_{\mathcal{C}_t}$ using the sparse representation error $\mathbf{E}_{\mathcal{C}_t}$:
\begin{align}
\label{eq:boh}
\mathbf{W}_{\mathcal{C}_t} = \text{diag}\left(\frac{1}{||\mathbf{e}_1||_2^2},...,\frac{1}{||\mathbf{e}_{N_r}||_2^2}\right).
\end{align}
\normalsize
We then solve the optimization problem
\begin{align}
\label{eq:DL_DOMINODLW}
\mathbf{\hat{D}}_{\mathcal{C}_t} = \underset{\mathbf{D}_{\mathcal{C}_t} \in \mathcal{D}}{\text{minimize}} \phantom{1} ||(\mathbf{Y}_{\mathcal{C}_t}-\mathbf{D}_{\mathcal{C}_t}\mathbf{X}_{\mathcal{C}_t})\mathbf{W}_{\mathcal{C}_t}^{\frac{1}{2}}||_{F}^2.
\end{align}
\normalsize
This leads to the weighted least squares solution
\begin{align}
\label{eq:mah}
\mathbf{\hat{D}}_{\mathcal{C}_t} = \mathbf{{Y}}_{\mathcal{C}_t} \mathbf{{W}}_{\mathcal{C}_t} \mathbf{Y}^T_{\mathcal{C}_t} (\mathbf{{Y}}_{\mathcal{C}_t} \mathbf{{W}}_{\mathcal{C}_t} \mathbf{Y^T}_{\mathcal{C}_t})^{-1}.
\end{align}
\normalsize
The dictionary $\mathbf{D}$ is then updated with the atoms $\mathbf{\hat{D}}_{\mathcal{C}_t}$ and its columns are normalized by their $\ell_2$-norms.

The $\mathbf{D}$ is next used for updating the sparse coding of $\mathbf{{Y}}_{\mathcal{C}_t}$ using batch OMP. Selecting a value for the maximal residual error $\delta$ in (\ref{eq:omp}) is usually not straightforward. This value can be related to the amount of noise in the observed data but this information is not known. The samples of our training set can be seen as realizations of a statistical process with an unknown distribution and therefore one can associate to these realizations the concept of \textit{statistical entropy}. 
We compute the normalized entropy of the mean vector of all the training set samples as
\begin{align}
\label{eq:entropy}
E(\bm{\mu}_{\mathbf{Y}})= - \sum_{i=1}^{M}P(\bm{\mu}_{\mathbf{y}_{\text{t }}})\log{P(\bm{\mu}_{\mathbf{y}_{\text{t } }})},
\end{align}
where $\bm{\mu}_{\mathbf{Y}}$ is the mean vector of all training samples, $M$  is the number of features for each training sample and $P(\cdot)$ is the probability mass function. In our case, $P(\cdot)$ is obtained as the normalized histogram of $\bm{\mu}_{\mathbf{Y}}$. Here, $E(\bm{\mu}_{\mathbf{Y}})$ is an indicator of the \textit{randomness} of the data due to noise. We use $E(\bm{\mu}_{\mathbf{Y}})$ as the maximal residual error $\delta$ while applying batch OMP in DOMINODL. 
Algorithm~\ref{alg:dominodl} summarizes all major steps of DOMINODL. 

\begin{algorithm}[ht]
\LinesNumbered
  \caption{Drop-Off MINi-Batch Online Dictionary Learning}
  \label{alg:dominodl}
  \SetAlgoLined
  \SetKwFor{Loop}{Loop}{}{EndLoop}
{
   \KwIn{ Training set ($\mathbf{Y}$), number of trained atoms ($K$), mini-batch dimension for new training data ($N_b$), mini-batch dimension for previous training data ($N_r$), drop-off value ($N_u$), convergence threshold ($\chi\in\mathbb{R}$)  }
   \KwOut{ Learned dictionary ($\mathbf{D}$), sparse decomposition of the training set ($\mathbf{X}$) }
   \BlankLine
   Generate initial dictionary $\mathbf{D}$ of dimension $K$ using training samples \par
   Normalize the columns of $\mathbf{Y}$ and $\mathbf{D}$ by their $\ell_2$-norms\par
   Sparsely decompose $\mathbf{Y}$ with the initial dictionary using entropy-thresholded batch OMP\par
   \Loop{}{
        Gather a mini-batch of new training set elements $\mathbf{Y}_{\mathcal{B}_t} =  \{\mathbf{y}_l: l \in \mathcal{B}_t\}$ with $\mathcal{B}_t = \{\mathbf{y}_l: i \le l < i+N_b\}$ and $|\mathcal{B}_t| = N_b$ \par
        SR of $\mathbf{Y}_{\mathcal{B}_t}$ with the dictionary $\mathbf{D}$ using entropy-thresholded batch OMP \par
        Store the set of atoms $\mathbf{D}_{\mathcal{B}_t}$ participating in the SR of $\mathbf{Y}_{\mathcal{B}_t}$ \par
        Randomly select a mini-batch of previous training set elements $\mathbf{Y}_{\mathcal{M}_t}= \{\mathbf{y}_l: l \in \mathcal{M}_t\}$ with  $|\mathcal{M}_t| = N_r$ and $\mathcal{M}_t \cap \mathcal{M}_{t-1} = \emptyset$ \par 
        Consider a subset $\mathbf{Y}_{\mathcal{A}_t} = \{\mathbf{y}_l: l \in \mathcal{A}_t\}$ where $\mathcal{A}_t \subset \mathcal{M}_t$ such that $\mathcal{A}_t = \left\lbrace l : l \in \mathcal{M}_t, \left\langle \mathbf{x}_l^T \cdot \mathbf{x}_t \right\rangle \neq 0 \right\rbrace$  \par
        Form $\mathbf{Y}_{\mathcal{C}_t} = \mathbf{Y}_{\mathcal{A}_t} \cup \mathbf{Y}_{\mathcal{B}_t}$ and store $\mathbf{D}_{\mathcal{C}_t}$ the atoms of $\mathbf{D}$ shared by $\mathcal{B}_t$ and $\mathcal{M}_t$\par

        Compute the errors $\mathbf{E}_{\mathcal{C}_t} = \mathbf{Y}_{\mathcal{C}_t} -\mathbf{D}_{\mathcal{C}_t}\mathbf{X}_{\mathcal{C}_t} = [\mathbf{e}_1, \cdots,\mathbf{e}_{N_r}]$ \par
        
        Form the weighting matrix $\mathbf{W}_{\mathcal{A}_t} = \text{diag}\left(\frac{1}{||\mathbf{e}_1||_2^2},...,\frac{1}{||\mathbf{e}_{N_r}||_2^2}\right)$ \par
        Update $\mathbf{\hat{D}}_{\mathcal{C}_t}$:  $\mathbf{\hat{D}}_{\mathcal{C}_t} = \mathbf{{Y}}_{\mathcal{C}_t} \mathbf{{W}}_{\mathcal{C}_t} \mathbf{Y}^T_{\mathcal{C}_t} (\mathbf{{Y}}_{\mathcal{C}_t} \mathbf{{W}}_{\mathcal{C}_t} \mathbf{Y}^T_{\mathcal{C}_t})^{-1}$ and normalize its columns \par
        Replace the updated atoms $\mathbf{D}_{\mathcal{C}_t}$ into $\mathbf{D}$ and normalize its columns \par
        Perform SR of selected signals used in the previous step using entropy-thresholded batch OMP \par
        Eliminate previous training set elements which have not been used for the last $N_u$ iterations  \par
       \textbf{if} { ${\vert\vert\left(\mathbf{Y}_{\mathcal{C}_t}- \mathbf{D}_{t}\mathbf{X}_{\mathcal{C}_t}\right) \left( \mathbf{W}_{i}\right)^{0.5} \vert\vert_F^2 < \chi}$ } \textbf{then} break
   }
}
\end{algorithm}

Table~\ref{tbl:dlcomparison} summarizes the important differences between DOMINODL and other related algorithms. 
Like MOD and CBWLSU, DOMINODL uses a weighted least squares solution in the dictionary update. The proof of convergence for the alternating minimization method in MOD was provided in \cite{agarwal2014learning} where it is shown that alternating minimization converges linearly as long as the following assumptions hold true: sparse coefficients have bounded values, sparsity level is on the order of $\mathcal{O}(M^{1/6})$ 
 and the dictionary satisfies the RIP property. In \cite{naderahmadian2016correlation}, these assumptions have been applied for CBWLSU convergence. Compared to CBWLSU, the improvements in DOMINODL include mini-batch based data selection and data reduction via a drop-off strategy but the update algorithms remain the same. Numerical experiments in Section~\ref{sec:exp} suggest that DOMINODL usually converges in far fewer iterations than CBWLSU. 

Although we developed and tested DOMINODL on a highly correlated GPR dataset (see Section \ref{sec:meas_camp}), this technique may be employed in other applications where real-time learning is necessary and the signals are correlated. Our tests demonstrate that DOMINODL converges faster than other online DL approaches (see Section \ref{sec:exp}) because of the combined strategy of drawing more new elements for each iteration, considering less previous elements in search for correlation and dropping off the unused previous elements. The entropy-based calculation of $\delta$, although not exclusive for DL applications as mentioned above, also helps in improving the SR of the data thus, learning a more representative dictionary.

\begin{figure}[t]
\centering
  \includegraphics[scale=0.19]{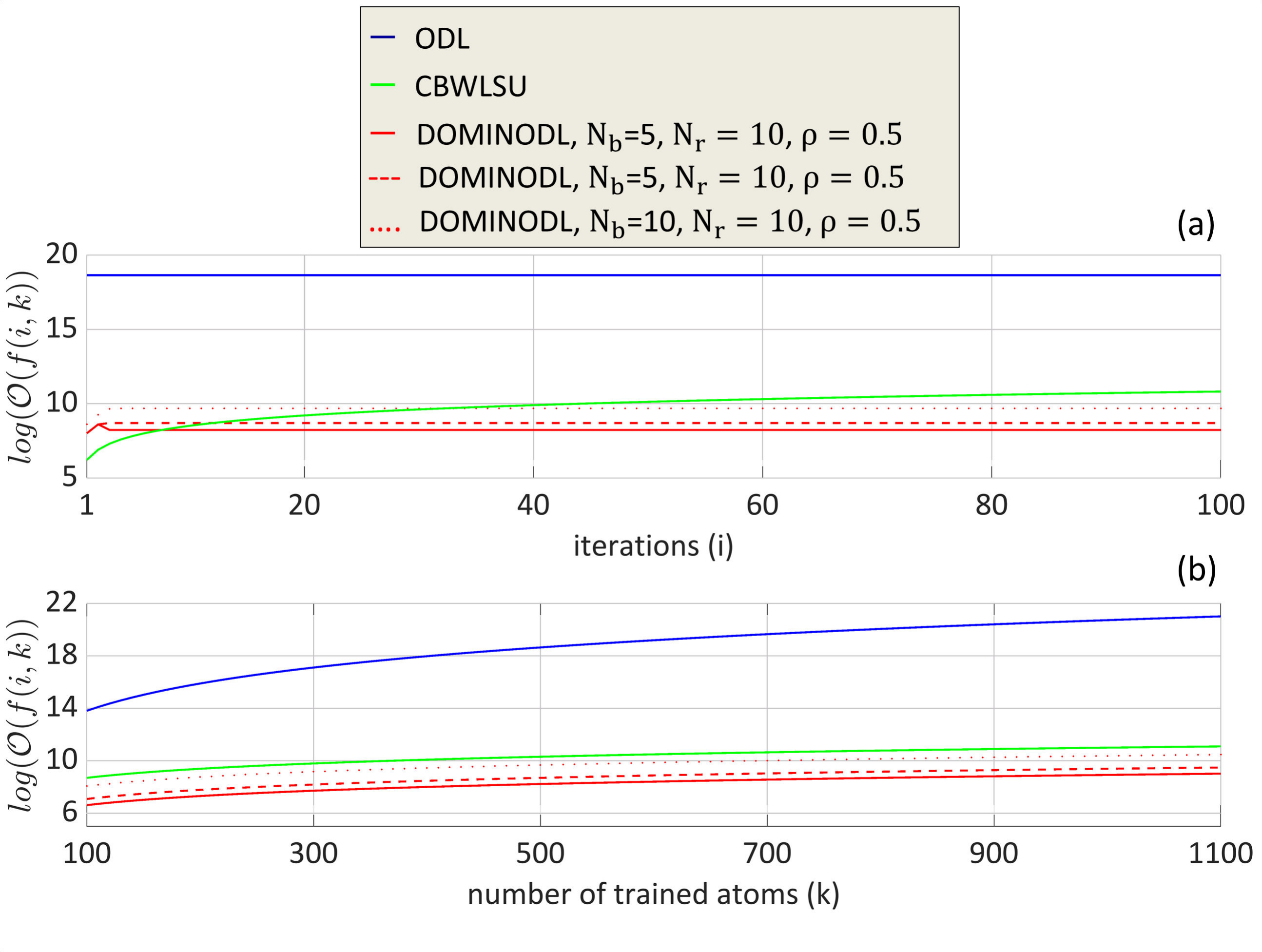}
  \caption{Computational complexity of online DL strategies for increasing number of (a) iterations and (b) trained atoms.}
\label{fig:compcomp}
\end{figure}

Computational complexity of DOMINODL is very low compared to other online approaches. As mentioned earlier, there are $K$ atoms in the dictionary. Assume that every signal is represented by a linear combination of $K_s$ atoms, $K_s \ll K$. Empirically, among all possible combinations of $K_s$ atoms from $K$, the probability to have a common atom in the sparse representation is $K_s/K$. Given $L$ training elements, the number of training data which have a specific atom in their representation is proportional to $LK_s/K$. Suppose our mini-batch has elements that reduce the number of training data by a factor $\beta < 1$ (depending on the values of $N_r$ and $N_b$). Further, assume that the dropping off step reduces the training set elements by a factor $\rho < 1$. The number of training data $L_i$ in the $t^{\text{th}}$ iteration is proportional to $\beta\rho iK_s/ K$. Then, the worst estimate of DOMINODL's computational complexity is due to the sparse coding batch OMP which is of order $\mathcal{O}(L_tK^2) = \mathcal{O}(\beta\rho tK_sK) \approx \mathcal{O}(\beta\rho tK)$. This is much smaller than the complexity of ODL ($=\mathcal{O}(K^3)$) or CBWLSU ($=\mathcal{O}(tK)$).\\
Figure~\ref{fig:compcomp} illustrates the computational complexity of online DL approaches. Figure~\ref{fig:compcomp}(a) shows that, for fixed number of iterations ($t=60$), the general trend of complexity with respect to the increase in the number of atoms ($K$) is similar for all algorithms. However, the complexity of ODL is higher than CBWLSU and DOMINODL; the latter being the least complex. When the number of iterations is increased, the complexity of ODL and CBWLSU have a similar increasing trend (see Fig.~\ref{fig:compcomp}(b)). In case of DOMINODL, its complexity is similar to the increasing trend of CBWLSU and determined largely by $N_b$. When DOMINODL iterations begin accounting for $N_r$ previous elements, its complexity stays constant. The value of $\beta$ changes for every iteration, while $\rho$ depends on the data itself. In general, after a few dozen of iterations, DOMINODL's complexity always stays lower than CBWLSU.
\section{Measurement Campaign}
\label{sec:meas_camp}
In this section, we first provide details of our GPR system and the field measurement campaign. We then describe the procedure to organize the entire dataset for our application.

\subsection{Ground Penetrating Radar System}
\label{subsec:GPR_syst}
\begin{figure}[t]
	\centering
	\includegraphics[width=0.35\columnwidth]{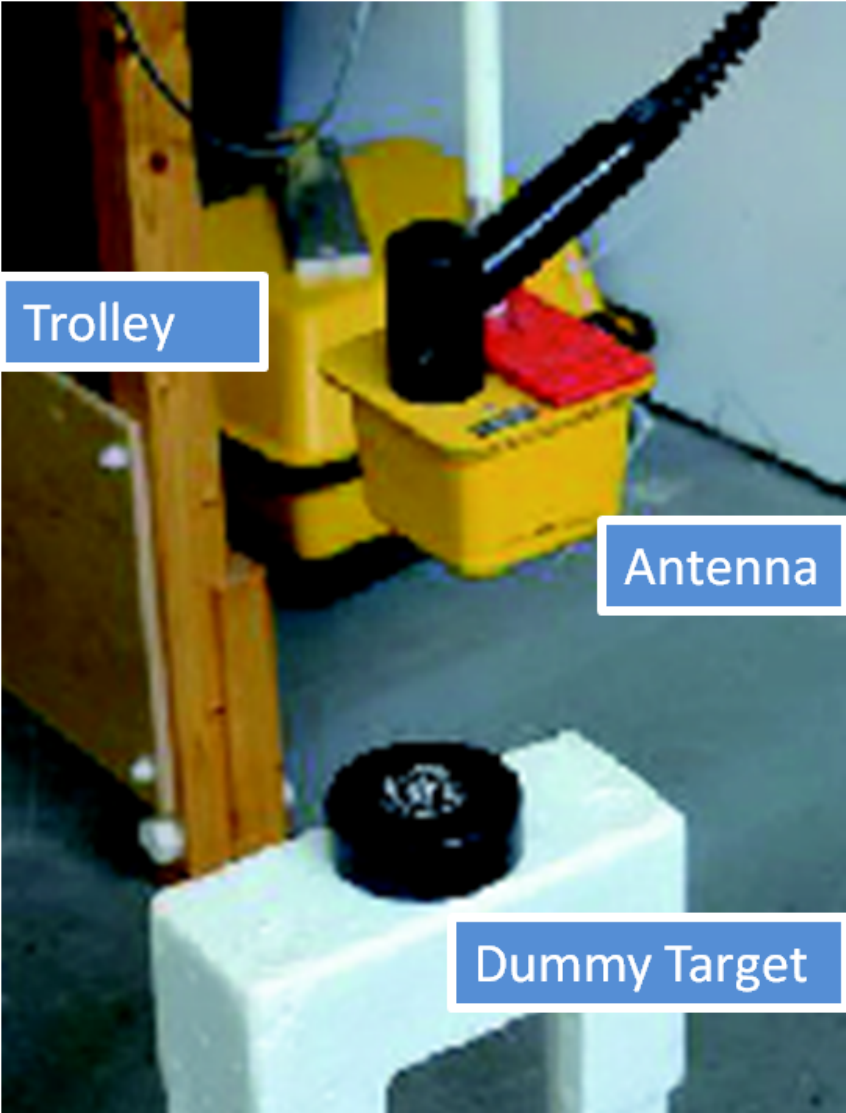}
  	\caption{The L-band GPR system is attached to a movable trolley platform. It is mounted along a rail system and scans the target from above.}
	\label{fig:GPR}
\end{figure}

Our GPR (see Fig. \ref{fig:GPR}) is the commercially available SPRScan system manufactured by ERA Technology. It is an L-band, impulse waveform, ultra-wideband (UWB) radar that is mounted on a movable trolley platform. Pulsed GPRs are more effective in terms of offering penetration depth and wide bandwidth with respect to the standard Stepped-Frequency Continuous Wave (SFCW) systems. The former is also more robust to electronic interference and does not suffer from unequal balancing of antenna signals \cite{leckebusch2011comparison}.
\begin{figure*}
\centering
  \includegraphics[scale=0.35]{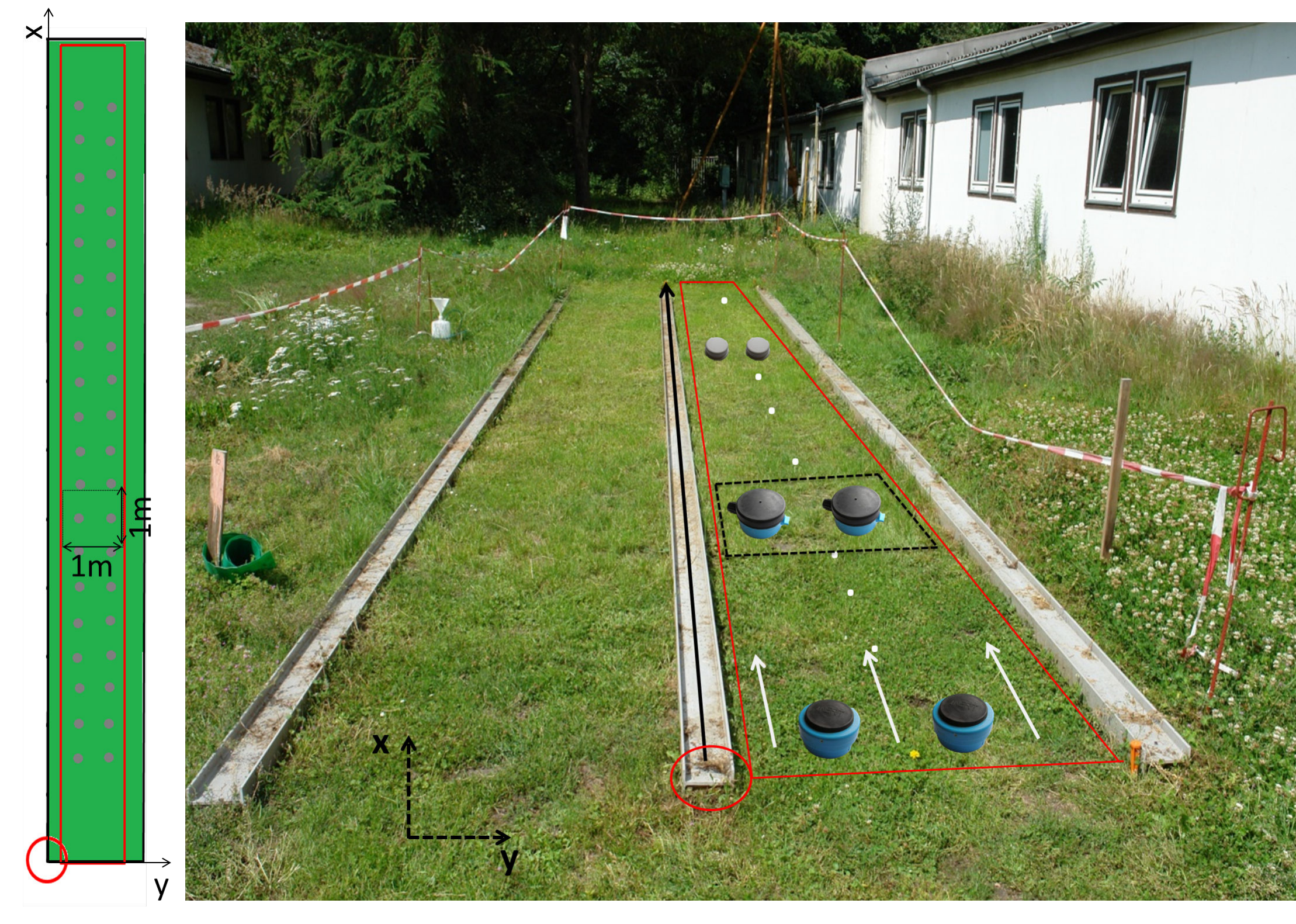}
  \caption{The LIAG test field in Hannover (right) along with its layout (left). The scan directions X and Y of the radar are indicated on the photograph and layout. The radar coverage region is indicated by solid red lines with a red circle showing the origin of the scan. The white arrows in the photograph indicate specific lanes scanned in the X direction that are separated in the Y direction by $4$ cm. In the layout, each gray dot represents the location of a buried test target. An individual survey area unit of $1$ m $\times$ $1$ m that contains 2 targets is also indicated on the layout (solid black lines) and the photograph (dotted black lines). The solid black arrow over the middle rail in the photograph is where the SPRScan was mounted.}
\label{fig:field}
\end{figure*}
\begin{figure}
\centering
  \includegraphics[width=0.75\columnwidth]{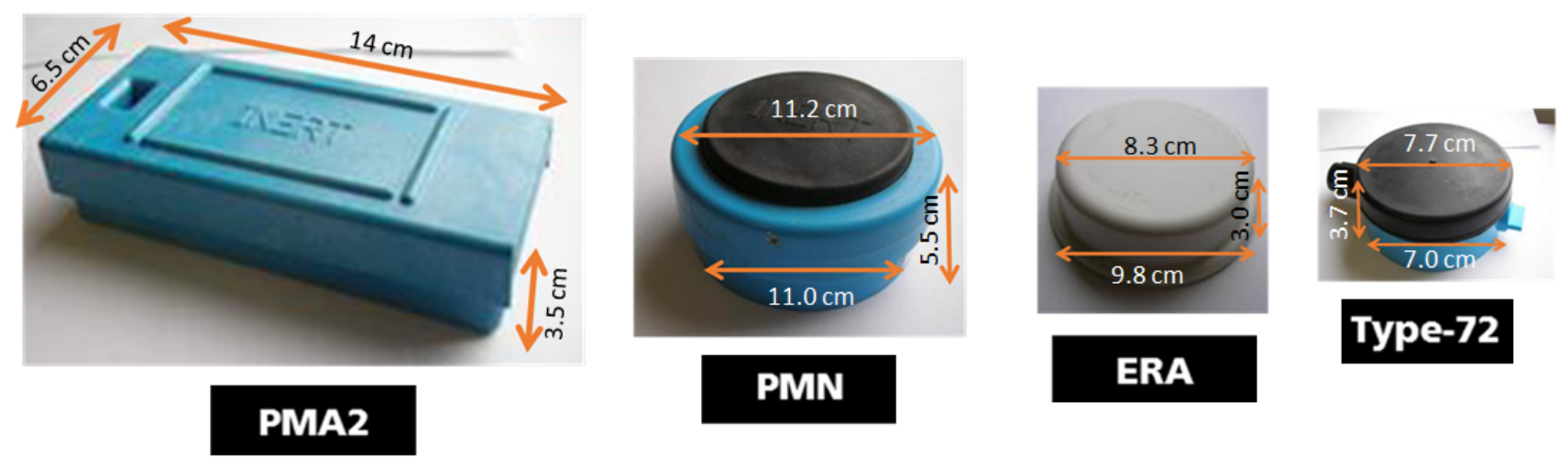}
  \caption{Details of the simulant landmines and the standard test target buried in the test field.}
\label{fig:mines}
\end{figure}
Table \ref{tbl:techparams} lists the salient technical parameters of the system. The radar uses a $8\times8$ cm dual bow-tie dipole antenna for both transmit (Tx) and receive (Rx) sealed in a metallic shielding filled with an internal absorber. The central frequency of the system ($f_c$) and its bandwidth ($\Delta f$) are $2$ GHz. 
The pulse repetition frequency (PRF) and the sampling of the receiver ADC is $1$ MHz. The scanning system has a resolution of $1$ cm towards the perpendicular broadside (or X direction) and $4$ cm towards the cross-beam (Y direction). In our field campaigns, the SPRScan system moves along the survey area over a rail system which allows accurate positioning of the sensor head in order to obtain the aforementioned resolution in X and Y (see also Section~\ref{sec:meas_camp}).
\begin{table}[t]
\centering
    \caption{Technical characteristics of impulse GPR}
	\label{tbl:techparams}
	\begin{tabular}{ p{4.0cm} | p{2.5cm} }
		\hline
         \noalign{\vskip 1pt}    
         	Parameter & Value\\[1pt]
		\hline
		\hline
        \noalign{\vskip 1pt}    
		   	Operating frequency & 2 GHz\\[1pt]
		   	Pulse repetition frequency & 1 MHz\\[1pt]
		   	Pulse length & 0.5 ns\\[1pt]            
            Sampling time & 25 ps\\[1pt]
            Spatial sampling along the beam & 1 cm\\[1pt]
            Cross-beam resolution & 4 cm\\[1pt]
            Antenna height & 5-9 cm\\[1pt]
            Antenna configuration & Perpendicular broadside\\[1pt]
            Samples/A-scan & 512\\
		\hline
		\hline
        \noalign{\vskip 2pt}
	\end{tabular}
\end{table}

The transmit pulse of the GPR system is a \textit{monocycle}. Given the Gaussian waveform
\begin{align}
\label{eq:Gauss}
s_G(t)= Ae^{-2\pi^2 f_c^2 (t-\tau)^2 },\;t\in[0,+\infty],
\end{align}
where $f_c$ is the central frequency, $A$ is the peak amplitude and $\tau=1/f_c$, the monocycle waveform is its first derivative \cite{warren2015advanced}
\begin{align}
\label{eq:Mono}
s_T(t)= -4 \pi^2 f_c^2 A (t-\tau) e^{-2\pi^2 f_c^2 (t-\tau)^2 },\;t\in[0,+\infty].
\end{align}
In these UWB systems both the central frequency and the bandwidth are approximately the reciprocal of the pulse length.

The scattering of UWB radar signals from complex targets that are composed of a finite number of scattering centers can be described in terms of the channel impulse response (CIR). Here, the CIR is considered as a linear, time invariant, causal system which is a function of the target shape, size, constituent materials, and scan angle. The CIR $h(t)$ of a GPR target, with $M$ scatterers, is expressed as a series of time-delayed and weighted Gaussian pulses \cite{hussain1990principles}
\begin{align}
h(t)= \sum\limits_{m=1}^{M}\alpha_m e^{-4\pi[(t-t_m)/\Delta T_m]^2},
\end{align}
where each scatterer located at range $r_m$ from the radar is characterized by the reflectivity $\alpha_m$, duration $\Delta T_m$, relative time shift $t_m = 2r_m/v_s$, where $v_s=c/\sqrt{\epsilon_r}$ is the speed of the electromagnetic wave in the soil, $c=3\times 10^8$ m/s is the speed of light, and $\epsilon_r$ is the dielectric constant which depends on the soil composition and moisture.

The response of the target to the Gaussian monocycle is the received signal
\begin{align}
\label{eq:rng_prof}
y(t)= s_T(t)\ast h(t),
\end{align}
also regarded as the target image, or \textit{range profile}. For each X/Y position, the system receives a radar echo (range profile) from the transmitted pulse. In order to deal with the exponential signal attenuation during the propagation through the soil medium, the dynamic range of the signal is enhanced via stroboscopic sampling \cite{daniels2005ground,pasculli2015real,bystrov2013analysis}. This technique comprises integrating $N$ receiver samples (generated by transmitting a sequence of $N$ pulses) at the ADC receiver sampling rate but with a small time offset $\delta$ for each of them. To achieve the desired stroboscopic sampling rate $T_s$, the time offset must be selected accordingly, i.e., $\delta= T_s/ N$ \cite{bystrov2013analysis}. Our GPR system employs stroboscopic sampling to reach a pseudo sampling frequency of $f_s=1/T_s=40$ GHz (much above the Nyquist rate) to yield the discrete-time signal $y[n] = y(nT_s)$.

The receiver has the ability to acquire a maximum of 195 profiles per second, each one consisting of 512 range samples. Prior to the A/D conversion, the signal is averaged to improve the signal to noise ratio (SNR). A time-varying gain correction can be applied to compensate for the soil attenuation and increase the overall dynamic range of the system.  The receiver averages 100 range profiles for each antenna position. 

\subsection{Test Field Measurements}
\label{subsec:test_field_meas}
We evaluated the proposed approach with the measurement data from a 2013 field campaign at Leibniz Institute for Applied Geophysics (LIAG) in Hannover (Germany) \cite{gonzalez2013combined}; Fig.~\ref{fig:field} shows the test field, for detailed ground truth informations. The soil texture was sandy and highly inhomogeneous (due to the presence of material such as organic matter and stones), thereby leading to a high variability in the electrical parameters. We measured the dielectric constant at three different locations of the testbed with a Time Domain Reflectometer (TDR) to obtain an estimate of its mean value and variability. The average value oscillated between 4.6 and 10.1 with $15\%$ standard deviation and \textit{correlation length} \cite{gonzalez2013combined} of $20$ cm. These large variations in soil dielectric characteristics pose difficulties in mine detection.

During the field tests, the SPRScan system moved on two plastic rails with the scan resolution in the X and Y directions being $1$ and $4$ cm, respectively. The entire survey lane was divided in $1\times1$ m sections (see Fig.~\ref{fig:field}), each containing two targets in the center. The targets on the left and right sides of the lane were buried at approximately $10$ and $15$ cm depths, respectively.

Our testbed contains standard test targets (STT) and simulant landmines (SIM) of different sizes and shapes. An STT is a surrogate target used for testing landmine detection equipment. It is intended to interact with the equipment in an identical manner as a real landmine does. An SIM has the representative characteristics of a specific landmine class although it is not a replica of any specific model. In this paper, we study three STTs (PMA2, PMN and Type-72) and one SIM(ERA). All of these test objects are buried at a depth of $10$-$15$ cm in the test field \cite{gonzalez2013accurate}. For classification purposes, we group PMN and PMA2 together as the largest targets while T72 mines are the smallest (Fig.~\ref{fig:mines}). 

\subsection{Dataset Organization}
\label{subsec:data_organization}
The entire LIAG dataset consists of 27 aforementioned survey sections (or simply, ``surveys'') of size $1\times1$ m. Every survey consists of $2500$ range profiles. We arranged the data into the training set ($\mathbf{Y}$) to be used for both DL and classification (as explained in subsection \ref{subsec:GPRCLASS}) and a test set ($\mathbf{Y}_{\text{TEST}}$) to evaluate the performance of the proposed algorithms. 

The training set $\mathbf{Y} \in \mathbb{R}^{M \times L}$ is a matrix whose $L$ columns $\{\mathbf{y}_i\}_{i=1}^L$ consist of sampled range profiles $\mathbf{y}_i = \big[y[0],\cdots,y[M-1]\big]^T$ of $M$ range profiles each. The profiles are selected from different surveys and contain almost exclusively either a particular class of landmine or clutter. In total, we have $463$, $168$, $167$ and $128$ range profiles for clutter, PMA2/PMN, ERA and Type-72, respectively. An accurate separation of these classes was very challenging because of the contributions from the non-homogeneous soil clutter that often masked the target responses completely. A poor selection would lead the DL to learn a dictionary that is appropriate for sparsely representing clutter, instead of landmines. The test set $\mathbf{Y}_{\text{TEST}} \in \mathbb{R}^{M \times J}$ is a matrix with $J=15000$ columns $\{\mathbf{y}_{\text{TEST}_i}\}_{i=1}^J$ that correspond to sampled range profiles from 6 surveys, two for each target class. The test and training sets contain data from separate surveys to enable fair assessment of the classification performance.

We denote by the matrices $\mathbf{X} \in \mathbb{R}^{K\times L}$ and $\mathbf{X_{\text{TEST}}} \in \mathbb{R}^{K\times J}$ the SRs of $\mathbf{Y}$ and $\mathbf{Y}_{\text{TEST}}$, respectively and $K$ by the number of atoms of the learned dictionary $\mathbf{D} \in \mathbb{R}^{M\times K}$.


\section{Parametric Analysis}
\label{sec:param}
In practice, the SR-based classification performance is sensitive to the input parameters of DL algorithms thereby making it difficult to directly apply DL with arbitrary parameter values. Previous works set these parameters through hit-and-trial or resorting to metrics that are unable to discriminate the influence of different parameters \cite{shao2013sparse}. In this section, we propose methods to investigate the effect of the various input parameters on the learning performance and then preset the parameter to \textit{optimal} values that yield the dictionary $\mathbf{D}$ (for each DL method) optimized to sparsely represent our GPR data, therefore improving the quality of the features for classification (i.e. the sparse coefficients).

Table \ref{tbl:inputparam} lists these parameters (see Section~\ref{sec:prop_meth}): number of iterations $N_t$, number of trained atoms $K$, and DOMINODL parameters $N_b$, $N_r$ and $N_u$. We applied K-SVD, LRSDL, ODL, CBWLSU and DOMINODL separately on the training set for different combination of parameter values. In order to compare the dictionaries obtained from various DL algorithms, we use a \textit{similarity measure} that quantifies the closeness of the original training set $\mathbf{Y}$ with the reconstructed set $\hat{\mathbf{Y}}$ obtained using the sparse coefficients of the learned dictionary $\mathbf{D}$. From these similarity values, empirical probability density functions (EPDFs) for any combination of parameter values are obtained; we evaluate these EPDFs using statistical metrics described in Section~\ref{subsec:stat_met}. These metrics efficiently characterize the similarity between $\mathbf{Y}$ and $\hat{\mathbf{Y}}$ and lead us to an optimal selection of various DL input parameters for our experimental GPR dataset.
\begin{table}[t]
\centering
 	\caption{DL parameters}
	\label{tbl:inputparam}
	\begin{tabular}{ l | l }
		\hline
         \noalign{\vskip 1pt}
           	DL algorithm & Input parameters\\
        \noalign{\vskip 1pt}
		\hline
        \hline
        \noalign{\vskip 1pt}
            K-SVD & $N_{t}$, $K$  \\[1pt]
            LRSDL & $N_{t}$, $K$ \\[1pt]
            ODL & $N_{t}$, $K$ \\[1pt]
            CBWLSU & $K$ \\[1pt]
            DOMINODL & $K$, $N_b$, $N_r$, $N_u$ \\[1pt]
		\hline
		\hline
        \noalign{\vskip 1pt}
	\end{tabular}
\end{table}

\subsection{Similarity Measure}
\label{subsec:sim_meas}
Consider the cross-correlation between the original training set vector $\mathbf{y}_i$ and its reconstruction $\hat{\mathbf{y}}_i$: $\mathbf{r}_{\mathbf{y}_i,\hat{\mathbf{y}}_i}[l] = \sum\limits_{n=-\infty}^{+\infty}\mathbf{y}_i[n]\hat{\mathbf{y}}_i[n+l]$. The normalized cross-correlation is defined as
\begin{align}
\overline{\mathbf{r}_{\mathbf{y}_i,\hat{\mathbf{y}}_i}}[l] = \frac{\mathbf{r}_{\mathbf{y}_i,\hat{\mathbf{y}}_i}[l]}{\sqrt{\mathbf{r}_{\mathbf{y}_i,\mathbf{y}_i}[0]\mathbf{r}_{\hat{\mathbf{y}}_i,\hat{\mathbf{y}}_i}[0]}}.
\end{align}
For the vector $\mathbf{y}_i$, we define the similarity measure $s_i$ as
\begin{align}
s_i = \textrm{max}|\overline{\mathbf{r}_{\mathbf{y}_i,\mathbf{\hat{y}}_i}(l)}|,
\end{align}
where a value of $s_i$ closer to unity demonstrates greater similarity of the reconstructed data with the original training set. We compute $\{s_i\}_{i=1}^{L}$ for all vectors $\{\mathbf{y}_i\}_{i=1}^L$, and then obtain the normalized histogram or empirical probability density function (EPDF) $p_{s_{DL}}$ of the similarity measure. Here, the subscript DL represents the algorithm used for learning $\mathbf{D}$ e.g. ``K'', ``O'', ``C'' and ``D'' for K-SVD, ODL, CBWLSU and DOMINODL, respectively. Various parameter combinations for a specific DL method result in a collection of EPDFs. For a given DL method, our goal is to compare the epdfs of similarity measure by varying these parameters, and arrive at the thresholds of parameter values after which the changes in $p_{s_{DL}}$ are only incremental.

\begin{figure}
\centering
  \includegraphics[scale=0.3]{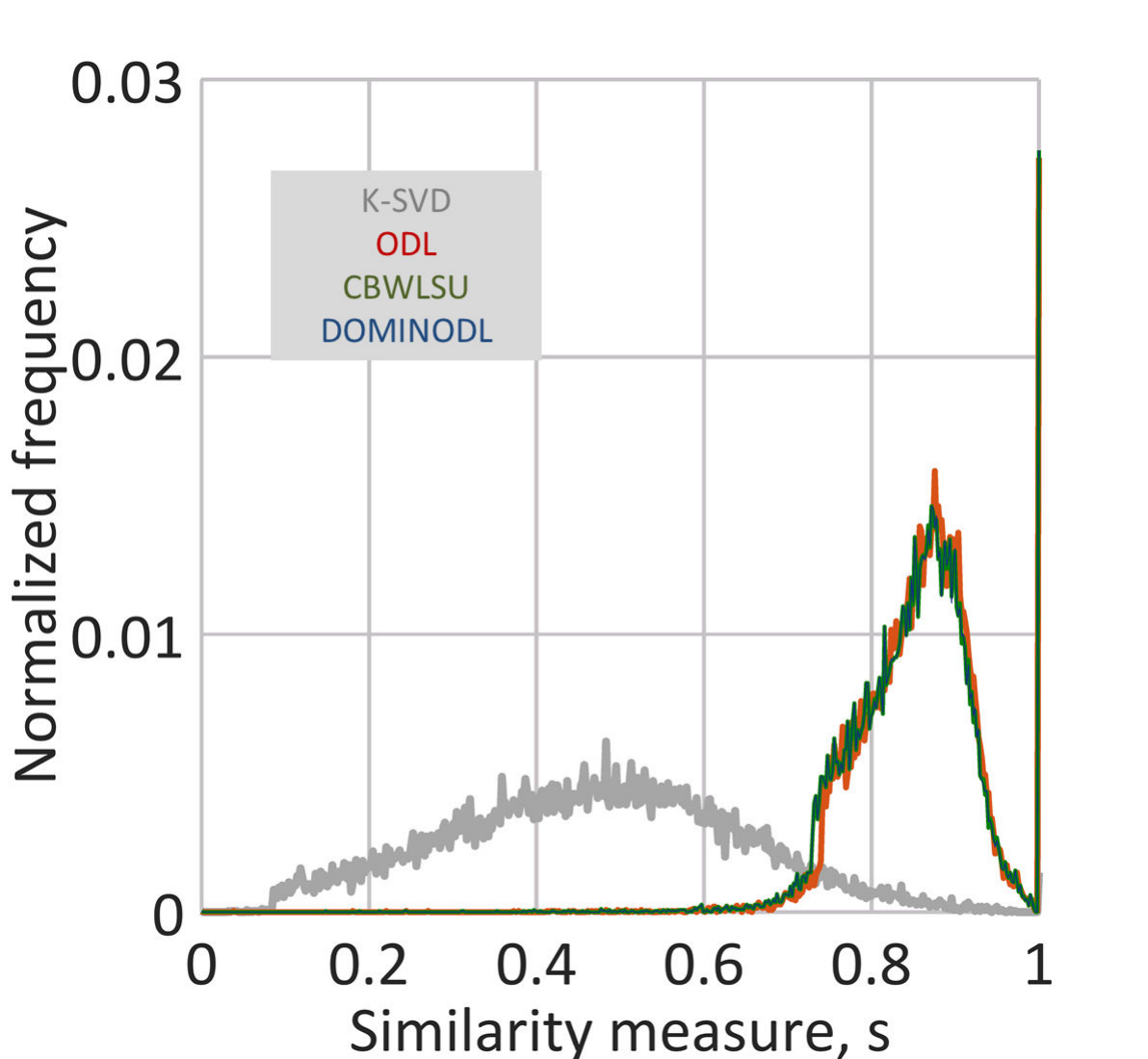}
  \caption{Normalized histograms of similarity measure using the following optimal parameters for the DL algorithm: $N_t=100$, $K=640$, $N_b=30$, $N_r=10$, and $N_u=10$. See Section~\ref{subsec:param_eval_dl} on the process to select these optimal values.}
\label{fig:odlKsvdDist}
\end{figure}

For instance, Fig.~\ref{fig:odlKsvdDist} shows the EPDFs of $s$ learned from the GPR mines data where optimal parameters for different DL methods were determined using statistical methods described in the following subsection. We note that the online DL approaches ($p_{s_{O}}$, $p_{s_{C}}$ and $p_{s_{D}}$) yield distributions that are more skewed towards unity than K-SVD ($p_{s_{K}}$). 
\subsection{Statistical Metrics}
\label{subsec:stat_met}
We are looking for parameter values for which $p_{s_{DL}}$ is skewed towards unity and has small variance. The individual comparisons of mean ($\mu$) and standard deviation ($\sigma$), as used in previous GPR DL studies \cite{shao2013sparse}, are not sufficient to quantify the observed dispersion in the epdfs obtained by varying any of the parameter values. Some DL studies \cite{shao2013sparse,hongxing2015interferometric,chen2016unsupervised} rely on bulk statistics such as NRMSE but these quantities are insensitive to large changes in parameter values and, therefore, unhelpful in fine-tuning the algorithms. For this evaluation, we will use three different metrics: the coefficient of variation, the Two-sample Kolmogorov-Smirnov (K-S) distance and the Dvoretzky-Kiefer-Wolfowitz (DKW) inequality.
\subsubsection{Coefficient of variation}
We choose to simultaneously compare both ($\mu$) and variance ($\sigma$) of a single EPDF by using the \textit{coefficient of variation}, $CV = \sigma/\mu$; in our analysis, it represents the extent of variability in relation to the mean of the similarity values.
\subsubsection{Two-sample Kolmogorov-Smirnov distance}
In the context of our application, it is more convenient to work with the cumulative distribution functions (CDFs) rather than with PDFs because the well-developed statistical inference theory allows for convenient comparison of CDFs. Therefore, our second metric to compare similarity measurements obtained by successive changes in parameter values is the \textit{two-sample Kolmogorov-Smirnov (K-S) distance} \cite{chakravarti1967handbook}, which is the maximum distance between two given empirical cumulative distribution functions (ECDF). Larger values of this metric indicate that samples are drawn from different underlying distributions. Given two random variables $s_1$ and $s_2$, suppose $\hat{F}_{s_1}$ and $\hat{G}_{s_2}$ are their ECDFs of the same length and correspond to their EPDFs $\hat{f}_{s_1}$ and $\hat{g}_{s_2}$, respectively. Then, the K-S distance is
\begin{align}
d_{ks}(\hat{F}_{s_1}, \hat{G}_{s_2}) = \sup_{1 \le i \le L}|\hat{F}_{s_1}(i) - \hat{G}_{s_2}(i)|,
\end{align}
where $\sup$ denotes the supremum over all distances and $L$ is the number of i.i.d. observations (or samples) to evaluate both distributions. 
In our case, $L$ is the number of range profiles in the training set. We first compute a reference ECDF ($\hat{G}_{s_{\text{ref}}}$) for each DL algorithm and fixed parameter values. For
our purposes, this reference ECDF will be obtained by a particular combination of input parameters of the selected DL algorithm. Then, we vary parameter values from this reference and obtain the corresponding ECDF $\hat{F}_{s_{\text{test}}}$ of similarity measure. Finally, we calculate the K-S distance $d_{ks}$ of $\hat{F}_{s_{\text{test}}}$ with respect to $\hat{G}_{s_{\text{ref}}}$ as
\begin{align}
d_{ks} = d_{ks}(\hat{F}_{s_{\text{test}}}, \hat{G}_{s_{\text{ref}}}) = \sup_{1 \le i \le L}|\hat{F}_{s_{\text{test}}}(i) - \hat{G}_{s_{\text{ref}}}(i)|.
\label{eq:kstest2}
\end{align}
For our evaluation, $d_{ks}$ states how much the selection of certain input parameters of DL changes the ECDFs of similarity values (i.e. how different is the result of DL) with respect to the reference. 
\subsubsection{Dvoretzky-Kiefer-Wolfowitz inequality metric}
As a third metric, we exploit the \textit{Dvoretzky-Kiefer-Wolfowitz inequality (DKW)} \cite{dvoretzky1956asymptotic,massart1990tight} which precisely characterizes the rate of convergence of an ECDF to a corresponding exact CDF (from which the empirical samples are drawn) for any finite number of samples. Let $d_{ks}(\hat{G}_s, F_s)$ be the K-S distance between ECDF $\hat{G}_s$ and the continuous CDF $F_s$ for a random variable $s$ and $L$ samples. Since $\hat{G}_s$ changes with the change in the $L$ random samples, $d_{ks}(\hat{G}_s, F_s)$ is also a random variable. We are interested in the
conditions that provide desired confidence in verifying if F and G are the same distributions for a given finite $L$. If the two distributions are indeed identical, then the DKW inequality bounds the probability that $d_{ks}$ is greater than any number $\epsilon$, with $0 < \epsilon < 1$ as follows\footnote{The corresponding asymptotic result that as $L\rightarrow \infty$, $d_{ks} \rightarrow 0$ with probability $1$ is due to the Glivenko-Cantelli theorem \cite{glivenko1933sulla,cantelli1933sulla}.}
\begin{align}
\label{eq:DKW}
\text{Pr}\left\lbrace d_{ks}\left( \hat{G}_s, F \right) > \epsilon \right\rbrace \le 2e^{-2L\epsilon^2}.
\end{align}
Consider a binary hypothesis testing framework where we use (\ref{eq:DKW}) to test the null hypothesis ${\mathcal{H}_0: F = \hat{G}}$ for a given ${\epsilon}$. The probability of rejecting the null hypothesis when it is true is called the p-value of the test and is bounded by the DKW inequality. Assuming the p-value is smaller than a certain confidence level $\alpha$, the following inequality must hold with probability at least $1-\alpha$:
\begin{align}
\label{eq:epsilon_DKW}
d_{ks}\left( \hat{G}_s, F \right) \leq \sqrt{-\frac{1}{2L}\text{ln}\left(\frac{\alpha}{2}\right)}.
\end{align}

Our goal is to use the DKW inequality to compare two ECDFs $\hat{F}_{s_{\text{test}}}$ and $\hat{G}_{s_{\text{ref}}}$ as in (\ref{eq:kstest2}), to verify if they are drawn from the same underlying CDF. By the triangle inequality, the K-S distance satisfies
\begin{align}
\label{eq:triangleinequ}
d_{ks}(\hat{F}_{s_{\text{test}}},\hat{G}_{s_{\text{ref}}}) = d_{ks}(\hat{F}_{s_{\text{test}}},{F}) + d_{ks}(\hat{G}_{s_{\text{ref}}},{F}),
\end{align} 
where $G$ an $F$ are the underlying CDFs corresponding to $\hat{G}$ and $\hat{F}$. We now bound the right side using DKW
\begin{align}
\label{eq:DKW2}
d_{ks}(\hat{F}_{s_{\text{test}}},\hat{G}_{s_{\text{ref}}}) &\leq \sqrt{-\frac{1}{2L}\text{ln}\left(\frac{\alpha}{2}\right)} + \sqrt{-\frac{1}{2L}\text{ln}\left(\frac{\alpha}{2}\right)}\nonumber\\
&=\sqrt{-\frac{2}{L}\text{ln}\left(\frac{\alpha}{2}\right)},
\end{align} 
which is the maximum distance for which $\hat{F}_{s_{\text{test}}}$ and $\hat{G}_{s_{\text{ref}}}$ are identical with probability $1-\alpha$. The \textit{DKW metric} is the difference
\begin{align}
\label{eq:DKW3}
d_{dkw} = \sqrt{-\frac{2}{L}\text{ln}\left(\frac{\alpha}{2}\right)} - 
d_{ks}(\hat{F}_{s_{\text{test}}},\hat{G}_{s_{\text{ref}}}) .
\end{align}
Larger values of this metric imply greater similarity betweem the two ECDFs; a negative value implies that the null hypothesis is not true.

\subsection{Parametric Evaluation}
\label{subsec:param_eval_dl}
We evaluated the performance of the aforementioned DL algorithms by analyzing the influence of the various DL input parameters using the metrics introduced in \ref{subsec:stat_met} for the reconstruction of the training set $\mathbf{Y}$. There are various soil types and scenarios for a landmine contaminated site. The LIAG test data provides an accurate representation of a practical scenario. Our metrics are general and derived from widely accepted statistical studies. Thus, their relevance to similar scenarios is very likely. As shown in table \ref{tbl:inputparam} the number of iterations $N_t$ is not relevant to CBWLSU and DOMINODL while the latter requires additional parameters to spacify the mini-batch dimensions and the iterations required to drop-off unused training set elements. We compute the K-S distance and the DKW metric for all methods with respect to a reference distribution $p_{\text{ref}}$. This reference, different for each DL algorithm, is obtained using the following parameters as applicable: $N_t = 1$, $K=300$, $N_b = 30$, $N_r=10$ and $N_u=10$. 


\begin{figure*}
\centering
  \includegraphics[scale=0.5]{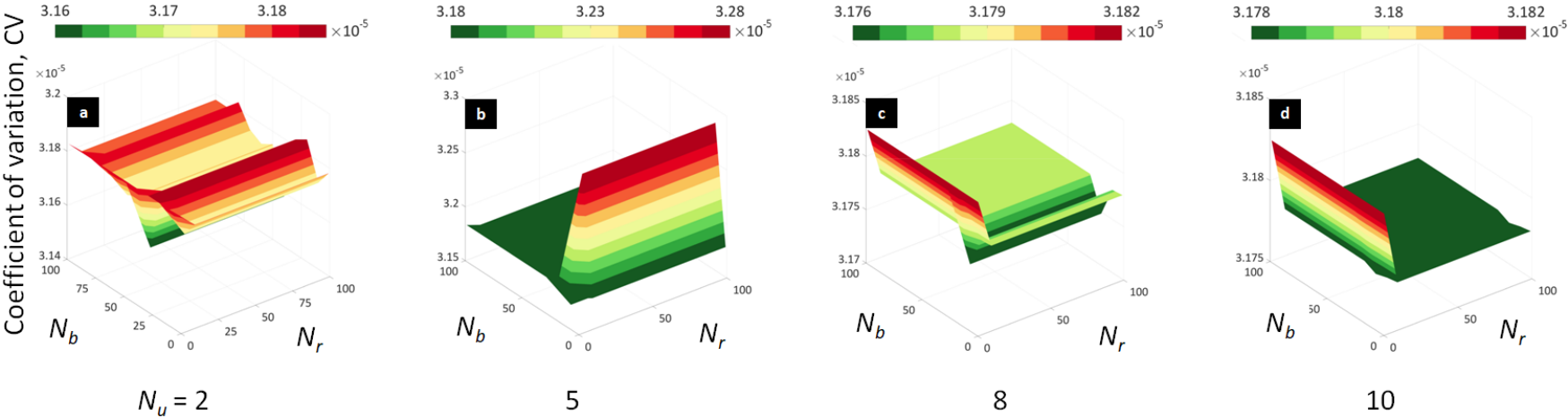}
  \caption{$CV$ as a function of DOMINODL input parameters for $k=640$ and $N_u$ as (a) 2, (b) 5, (c) 8, and (d) 10.}
\label{fig:DOMINOPAR}
\end{figure*}
\begin{figure}
\centering
  \includegraphics[scale=0.28]{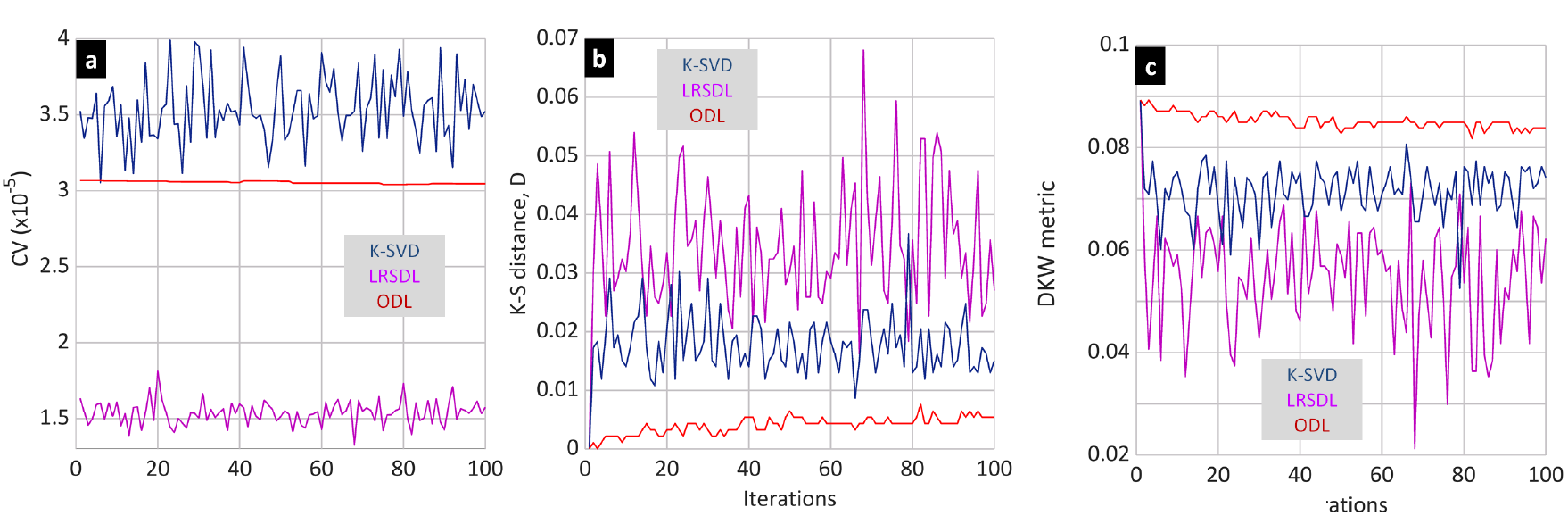}
  \caption{(a) $CV$, (b) K-S distance, and (c) DKW metric for K-SVD, LRSDL, and ODL parameter analyses as a function of the number of iterations $N_t$.}
\label{fig:cv_nit}
\end{figure}
\begin{figure}
\centering
  \includegraphics[scale=0.30]{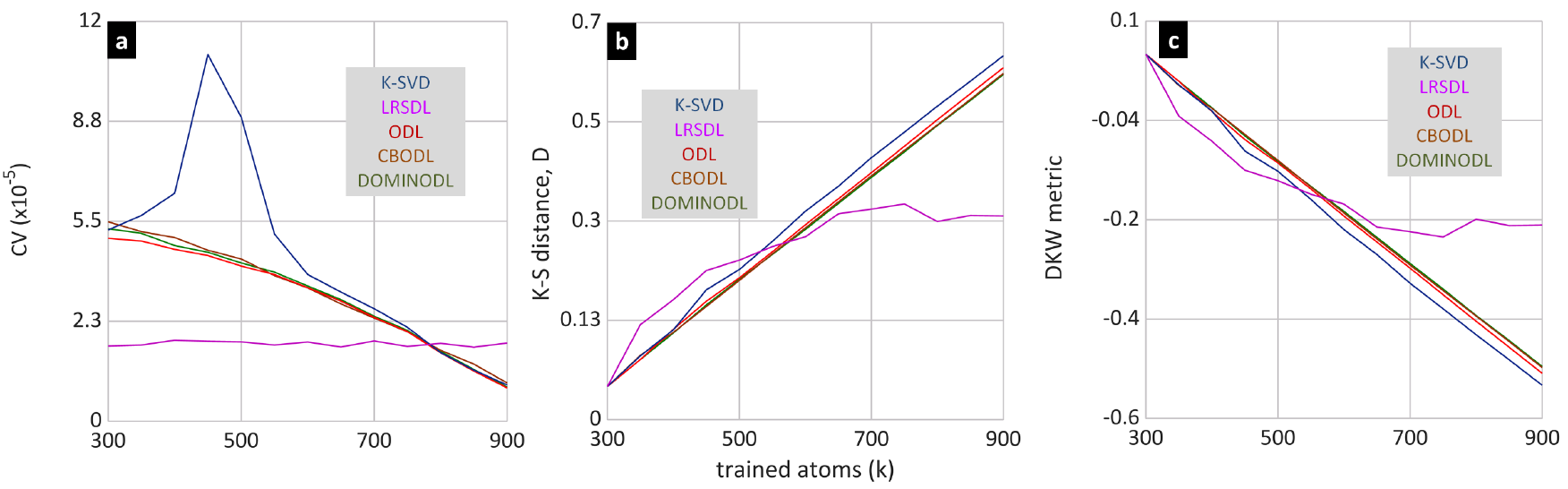}
  \caption{(a) $CV$, (b) K-S distance, and (c) DKW metric for various DL algorithms as a function of the number of trained atoms $K$.}
\label{fig:cv_k}
\end{figure}

\subsubsection{
Number of iterations}
Figures~\ref{fig:cv_nit}(a)-(c) show the effect of $N_t$ on the $CV$, K-S test distance $d_{ks}$ and the DKW metric $d_{W_L}$ for K-SVD, ODL and LRSDL. We have skipped CBWLSU and DOMINODL from this analysis because they do not accept $N_t$ as an input. For ODL, the $CV$ remains relatively unchanged with an increase in $N_t$. However, the K-SVD $CV$ exhibits an oscillating behavior and generally high values. In case of the K-S distance, ODL shows slight increase in $d_{ks}$ while K-SVD oscillates around a mean value that is higher than ODL. The DKW metric provides better insight: even though the ODL distributions differ from $p_{\text{ref}}$ with increase in the iterations, the null hypothesis always holds because $d_{dkw}$ remains positive. The $d_{dkw}$ for K-SVD is also positive but much smaller than ODL. It also does not exhibit any specific trend with an increase in iterations. We also observed a similar behavior with the mean of similarity values. The influence of the number of iterations in LRSDL had the same oscillating behaviour as in K-SVD but with larger variation. We conclude that the number of iterations $N_t$ does not significantly influence the metrics for these algorithms, and choose $N_t=100$.
\subsubsection{
Number of trained atoms}
Figs.~\ref{fig:cv_k}(a)-(c) compare all three metrics with change in the number of trained atoms $K$, a parameter that is common to all DL methods. We observe that $CV$ generally decreases with an increase in $K$. This indicates an improvement in the similarity between the reconstructed and the original training set. K-SVD shows an anomalous pattern for lower values of $K$ but later converges to a trend that is identical to other DL approaches. The K-S distance exhibits a linear change in the the distributions with respect to the reference. Since $d_{ks}$ quantifies the difference between the distributions rather than stating which one is better, combining its behavior with $CV$ makes it evident that an increase in $K$ leads to better distributions of similarity values. The DKW metric $d_{dkw}$, calculated with the same reference, expectedly also shows a linear change. It is clear that, even a slight change in $K$ leads to more negative values of $d_{dkw}$ implying that the null hypothesis does not hold true. This shows the significant influence of the parameter $K$ on the distributions. It was interesting to see a slight improvement for the coefficient of variation when using LRSDL with respect to the other strategies. However, KS-distance and DKW metric indicated that the distributions of similarity values for LRSDL were sensitive to the number of trained atoms only up to a certain value.
\subsubsection{DOMINODL 
parameters
}
It is difficult to evaluate DOMINODL EPDFs by varying all four parameters together. Instead, we fix the parameter that is common to all algorithms, i.e. the number of trained atoms $K$, and then determine optimal values of $N_b$, $N_r$ and $N_u$. Figure~\ref{fig:DOMINOPAR} shows the coefficient of variation $CV$ of the distribution of similarity values as a function of DOMINODL parameters. The drop-off value $N_u$ appears to have a greater influence than mini-batch dimensions $N_b$ and $N_r$. Our analysis of the computational times of DOMINODL showed that it is essentially independent of $N_r$ and $N_u$ but slightly increases with $N_b$. This is expected because we also increased the number of steps for sparse decomposition (see Algorithm~\ref{alg:dominodl}) which is the source of bulk of computations in DL algorithms \cite{naderahmadian2016correlation}. Further, in order to ensure that the correlation and the drop-off steps kick off from the very first iteration, DOMINODL should admit several new samples for each iteration thereby increasing $N_b$ as well as the number of previous elements accordingly. Taking into account these observations, we choose $N_b=30$ $N_r=10$ and $N_u=10$.

According to the results of the parametric evaluation, we choose the following combination of ``optimal'' parameters for testing our DL strategies:  $N_t=100$, $K=640$, $N_b=30$, $N_r=10$, and $N_u=10$.

\section{Experimental Results and Analysis} 
\label{sec:exp}
After selecting the input parameters of the proposed DL strategies, we proceed with the trained dictionaries for sparse decomposition of both training and test sets. The resulting sets of sparse coefficients are the input to the SVM classifier. As mentioned in Section~\ref{subsec:GPRCLASS}, the threshold $C$ and the kernel function parameter $\gamma$ for SVM have been selected through cross validation. Our key objective is to demonstrate that online DL algorithms may lead to an improvement in the classification performance over batch learning strategies. In particular, we want to analyze the performance of DOMINODL in terms of classification accuracy and learning speed. As a comparison with a popular state-of-the-art classification method, we also show the classification results with a deep-learning approach based on CNN. Finally, We demonstrate classification performance when the original samples of the range profiles are randomly reduced.
\begin{figure*}
\centering
  \includegraphics[width=0.8\textwidth]{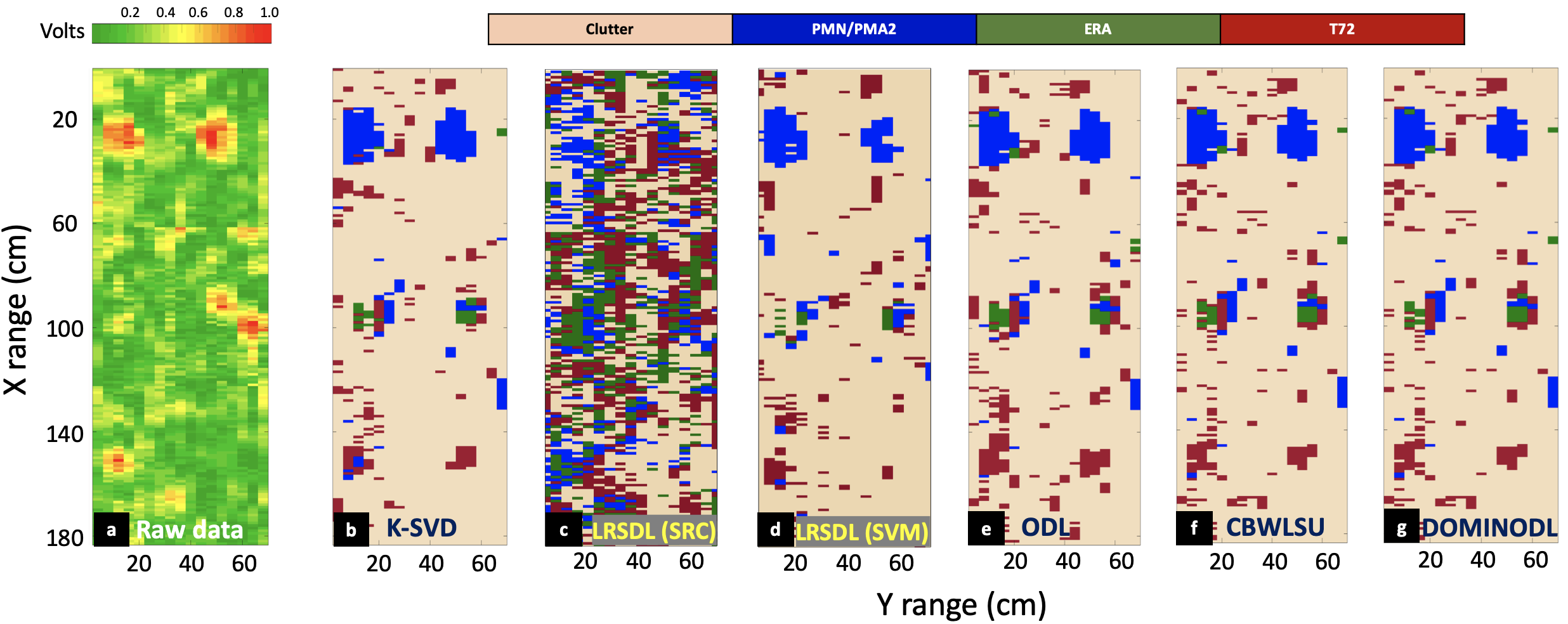}
  \caption{(a) Raw data at $15$ cm depth. The classification maps of the same area containing 6 buried landmines using an SR-based approach with dictionary learned using (b) K-SVD, (c) LRSDL (SRC), (d) LRSDL (SVM), (e) ODL, (f) CBWLSU, and (g) DOMINODL algorithms and optimally selected input parameters.}
\label{fig:CMAP_full}
\end{figure*}
\subsection{Classification with Optimal Parameters}
\label{subsec:class_opt}
For a comprehensive analysis of the classification performance, we provide both classification maps and confusion matrices for the test set $\mathbf{Y}_{\text{TEST}}$ using the optimal DL input parameters that we selected following our parametric evaluation in Section~\ref{sec:param}. The classification maps depict the predicted class of each range profile of the survey under test. The pixel dimension of these maps is dictated by the sampling of the GPR in X and Y directions (see Table~\ref{tbl:techparams}). We stitched together 3 of the 6 surveys from the test set $\mathbf{Y}_{\text{TEST}}$ where each survey had 2 buried landmines from a specific target class (PMN/PMA2, ERA and Type-72). 

Figure~\ref{fig:CMAP_full} shows the classification maps for different DL methods along with the raw data at depth $15$ cm. The selected survey area covers a total of $2880$ range profiles. The raw data in Fig.~\ref{fig:CMAP_full}(a) shows that only four of the six mines exhibit a strong reflectivity while the other two mines have echoes so weak that they are not clearly visible in the raw data. Figures~\ref{fig:CMAP_full}(b)-(d) show the results of the SR-based classification approaches using DL. All methods clearly detect and correctly classify the large PMN/PMA2 mines. In case of the medium-size ERA, the echoes are certainly detected as non-clutter but some of its constituent pixels are incorrectly classified as another mine. It is remarkable that the left ERA mine is recognized by our method even though it cannot be discerned visually in the raw data. Most of the false alarms in the map belong to the smallest Type-72 mines. This is expected because their small sizes produce echoes very similar to the ground clutter. On the other hand, when T-72 is the ground truth, it is correctly identified.

Using accurate ground truth information, we defined \textit{target halos} as the boundaries of the buried landmines. The dimension of the target halos varied depending on the mine size. Let the number of pixels and the declared mine pixels inside the target halo be $n_t$ and $n_m$, respectively. Similarly, we denote the number of true and declared clutter pixels outside the target halo by $n_c$ and $n_d$, respectively. Then, the probabilities of correct classification ($P_{CC}$) for each target class and clutter are, respectively,
\begin{align}
P_{CC_{\text{mines}}} = \frac{n_m}{n_t},\:\:\textrm{and}\:\:P_{CC_{\text{clutter}}} = \frac{n_d}{n_c}.
\end{align}
The $P_{CC}$ being the output of a classifier should not be mistaken as the radar's probability of detection $P_d$ which is the result of a detector. A detector declares the presence of a mine when only a few pixels inside the halo have been declared as mine; $P_{CC}$ provides a fairer and more accurate evaluation of the classification result. This per-pixel information can be easily used to improve the final detection result. For instance, the operator could set a threshold for the minimum number of pixels to be detected in a cluster so that a circle with center at the cluster centroid could be used as the detected mine. However, such a circle may exclude some of the mine pixels leading to a potential field danger. The per-pixel classification is then employed to determine the guard area around the mine circle.

A confusion matrix is a quantitative representation of the classifier performance. The matrix lists the probability of classifying the ground truth as a particular class. The classes listed column-wise in the confusion matrix are the ground truths while the row-wise classes are their predicted labels. Therefore, the diagonal of the matrix is the $P_{CC}$ while off-diagonal elements are probabilities of misclassification. 
\begin{table}[t]
\centering
	\begin{threeparttable}[b]
 		\caption{Confusion matrix with optimal DL input parameter selection}
		\label{tbl:CM_DL_full}
		\begin{tabular}{ c | l | l | l | l | l}
			\hline
    		\noalign{\vskip 1pt}
            \multicolumn{2}{c|}{} & Clutter & PMN/PMA2 & ERA & Type-72\\[1pt]
        	\hline
        	\hline
        	\noalign{\vskip 1pt}
             						& Clutter & \cellcolor{black!25}0.892& 0.044 & 0.25 & 0.37 \\[1pt]
             \multirow{2}{*}{K-SVD} & PMN/PMA2 & 0.022 & \cellcolor{black!25}0.938\tnote{1} & 0.166 & 0.074 \\[1pt]
             						& ERA & 0.021 & 0.017 & \cellcolor{black!25}0.472 & 0.018 \\[1pt]
             						& Type-72 & 0.064 & 0 & 0.111 & \cellcolor{black!25}0.537 \\[1pt]
			\hline
		    \noalign{\vskip 1pt}
            						& Clutter & \cellcolor{black!25}0.435 & 0.061 & 0.111 & 0.351 \\[1pt]
            \multirow{2}{*}{LRSDL (SRC)} & PMN/PMA2 & 0.155 & \cellcolor{black!25}0.289 & 0.319 & 0.259 \\[1pt]
            						& ERA & 0.172 & 0.372 & \cellcolor{black!25}0.361 & 0.278 \\[1pt]
            						& Type-72 & 0.237 & 0.272 & 0.208 & \cellcolor{black!25}0.111 \\[1pt]
 			\hline
		    \noalign{\vskip 1pt}
            						& Clutter & \cellcolor{black!25}0.889 & 0.114 & 0.333 & 0.463 \\[1pt]
            \multirow{2}{*}{LRSDL (SVM)} & PMN/PMA2 & 0.026 & \cellcolor{black!25}0.877 & 0.186 & 0.185 \\[1pt]
            						& ERA & 0.027 & 0 & \cellcolor{black!25}0.444 & 0.926 \\[1pt]
            						& Type-72 & 0.058 & 0.008 & 0.041 & \cellcolor{black!25}0.426 \\[1pt]
 			\hline
        	\noalign{\vskip 1pt}
            					   	& Clutter & \cellcolor{black!25}0.871 & 0 & 0.194 & 0.333 \\[1pt]
            \multirow{2}{*}{ODL} 	& PMN/PMA2 & 0.022 & \cellcolor{black!25}0.973 & 0.139 & 0 \\[1pt]
            					   	& ERA & 0.018 & 0.026 & \cellcolor{black!25}0.583 & 0.018 \\[1pt]
            					   	& Type-72 & 0.088 & 0 & 0.083 & \cellcolor{black!25}0.648 \\[1pt]
        	\hline
        	\noalign{\vskip 1pt}
           							& Clutter & \cellcolor{black!25}0.872 & 0.017 & 0.181 & 0.314 \\[1pt]
            \multirow{2}{*}{CBWLSU} & PMN/PMA2 & 0.023 & \cellcolor{black!25}0.973 & 0.153 & 0 \\[1pt]
            						& ERA & 0.025 & 0.008 & \cellcolor{black!25}0.528 & 0\\[1pt]
            						& Type-72 & 0.08 & 0 & 0.138 & \cellcolor{black!25}0.685 \\[1pt]
			\hline
	    	\noalign{\vskip 1pt}
            						& Clutter & \cellcolor{black!25}0.876 & 0.017 & 0.167 & 0.315 \\[1pt]
            \multirow{2}{*}{DOMINODL} & PMN/PMA2 & 0.023 & \cellcolor{black!25}0.974 & 0.138 & 0 \\[1pt]
            						& ERA & 0.027 & 0.008 & \cellcolor{black!25}0.58 & 0 \\[1pt]
            						& Type-72 & 0.077 & 0 & 0.11 & \cellcolor{black!25}0.685 \\[1pt]
 			\hline
			\hline
        	\noalign{\vskip 1pt}
		\end{tabular}
       	\begin{tablenotes}
			\item[1] Gray denotes the $\mathbf{P_{CC}}$ value for a specified class and DL algorithm
		\end{tablenotes}
    \end{threeparttable}
\end{table}    

For the classification map of Fig.~\ref{fig:CMAP_full}, Table~\ref{tbl:CM_DL_full} shows the corresponding confusion matrices for each DL-based classification approach. In general, we observe an excellent classification of PMN/PMA2 landmines (\texttildelow$98$\%), implying that almost every range profile in the test set which belongs to this class is correctly labeled. The $P_{cc}$ for the clutter is also quite high (\texttildelow$90$\%). This can also be concluded from the classification maps where the false alarms within the actual clutter regions are very sparse (i.e. they do not form a cluster) and, therefore, unlikely to be interpreted as an extended target. As noted previously, most of the clutter misclassification is associated with the Type-72 class. The ERA test targets show some difficulty with correct classification. But most of the pixels within its target halo are declared at least as some type of mine (which is quite useful in terms of issuing safety warnings in the specific field area). This result can be explained by the fact that ERA test targets do not represent a specific mine but have general characteristics common to most landmines. The Type-72 mines exhibit a $P_{cc}$  which is slightly higher with respect to ERA targets. This is a remarkable result because Type-72 targets were expected to be the most challenging to classify due to their small size.

Conventionally, as mentioned in \cite{vu2017fast}, LRSDL is used with a sparse-representation-based classification (SRC). However, applying this approach to our problem resulted in very low accuracy (an average of \texttildelow$20$\% across all classes as evident from Table~\ref{tbl:CM_DL_full}) and semi-random classification maps (Fig.~\ref{fig:CMAP_full}). This can be explained by the extreme similarity between the training set examples of different classes; mines and clutter are only slightly dissimilar in their responses and mine responses are generally hidden in the ground reflections. Each learned ``block'' $D_{c}$ differed only slightly from the other 
and, therefore, poor classification results are achieved with this dataset. On the other hand, when we used the dictionary learned with LRSDL with our SVM-based technique, we obtained better classification accuracy (see Table~\ref{tbl:CM_DL_full} and Fig.~\ref{fig:CMAP_full}). However, this performance is still inferior to K-SVD and, hence, even worse than the other online DL approaches.

All DL algorithms used for our sparse classification approach show very similar results for the clutter and PMN/PMA2 classes. However, online DL methods show higher $P_{CC}$ for the ERA and Type-72 targets than K-SVD. From Table~\ref{tbl:CM_DL_full}, the detection enhancement using the best of the online DL algorithms for PMN/PMA2 over K-SVD is $((0.974-0.938)\times100)/0.938 \approx4$\%. The improvements for ERA and T-72 are computed similarly as $23$\% and $28$\%, respectively. 

\subsection{Classification with Non-Optimal Parameters}
\label{subsec:class_nonopt}
In order to demonstrate how the quality of the learned dictionary affects the final classification, we now show the confusion matrices for a non-optimal selection of input parameters in different DL algorithms. Our goal is to emphasize the importance of learning a good dictionary by selecting the optimal parameters rather than specifying how each parameter affects the final classification result. We arbitrarily selected the number of trained atoms $K$ to be only $300$ for all DL approaches, reduce the number of iterations to $25$ for ODL and KSVD and, for DOMINODL, we use $N_r$=30, $N_b$=5 and $N_u$=2. Table~\ref{tbl:CM_DL_bad} shows the resulting confusion matrix. While the clutter classification accuracy is almost the same as in Table~\ref{tbl:CM_DL_full}, the $P_{cc}$ for PMN/PMA2 landmines decreased by \texttildelow$10$\% for most of the algorithms except ODL where it remains unchanged. The classification accuracy for ERA and Type-72 mines is only slightly worse for online DL approaches. However, in the case of K-SVD, the $P_{CC}$ reduces by \texttildelow$30$\% and \texttildelow$10$\% for ERA and Type-72, respectively. Clearly, the reconstruction and correct classification of range profiles using batch algorithms such as K-SVD is strongly affected by a non-optimal choice of DL input parameters. As discussed earlier in Section~\ref{subsec:param_eval_dl}, this degradation is likely due to the influence of $K$ rather than $N_t$.
\begin{table}[t]
\centering
 	\caption{Confusion matrix with non-optimal DL input parameter selection}
	\label{tbl:CM_DL_bad}
		\begin{tabular}{ c | l | l | l | l | l}
		\hline
    	\noalign{\vskip 1pt}
            \multicolumn{2}{c|}{} & Clutter & PMN/PMA2 & ERA & Type-72\\[1pt]
        \hline
        \hline
        \noalign{\vskip 1pt}
             						& Clutter & \cellcolor{black!25}0.853& 0.07 & 0.305 & 0.222 \\[1pt]
            \multirow{2}{*}{K-SVD} 	& PMN/PMA2 & 0.037 & \cellcolor{black!25}0.851 & 0.222 & 0.111 \\[1pt]
            						& ERA & 0.032 & 0 & \cellcolor{black!25}0.194 & 0.241 \\[1pt]
            						& Type-72 & 0.077 & 0.078 & 0.277 & \cellcolor{black!25}0.426 \\[1pt]
		\hline
        \noalign{\vskip 1pt}
            						& Clutter & \cellcolor{black!25}0.86 & 0.017 & 0.181 & 0.444\\[1pt]
            \multirow{2}{*}{ODL} 	& PMN/PMA2 & 0.016 & \cellcolor{black!25}0.973 & 0.097 & 0 \\[1pt]
            						& ERA & 0.022 & 0.008 & \cellcolor{black!25}0.638 & 0 \\[1pt]
            						& Type-72 & 0.1 & 0 & 0.083 & \cellcolor{black!25}0.555 \\[1pt]
        \hline
        \noalign{\vskip 1pt}
            						& Clutter & \cellcolor{black!25}0.887 & 0.078 & 0.319 & 0.352 \\[1pt]
            \multirow{2}{*}{CBWLSU} & PMN/PMA2 & 0.019 & \cellcolor{black!25}0.877 & 0.097 & 0 \\[1pt]
            						& ERA & 0.018 & 0.043 & \cellcolor{black!25}0.541 & 0\\[1pt]
            						& Type-72 & 0.074 & 0 & 0.042 & \cellcolor{black!25}0.648 \\[1pt]
		\hline
         \noalign{\vskip 1pt}
            						& Clutter & \cellcolor{black!25}0.888 & 0.078 & 0.319 & 0.352 \\[1pt]
            \multirow{2}{*}{DOMINODL} & PMN/PMA2 & 0.019 & \cellcolor{black!25}0.877 & 0.097 & 0 \\[1pt]
            						& ERA & 0.018 & 0.043 & \cellcolor{black!25}0.54 & 0 \\[1pt]
            						& Type-72 & 0.074 & 0 & 0.042 & \cellcolor{black!25}0.648 \\[1pt]
 		\hline
		\hline
        \noalign{\vskip 1pt}
	\end{tabular}
\end{table}   
\begin{table}[t]
\centering
	\begin{threeparttable}[b]
    	\caption{Computational times for DL algorithms}
		\label{tbl:times}
		\begin{tabular}{ l | l | l | l | l |l}
			\hline
         	\noalign{\vskip 1pt}
            	& DOMINODL & CBWLSU & ODL & K-SVD &LRSDL\\[1pt]
			\hline
        	\hline
        	\noalign{\vskip 1pt}
           		Time (seconds) & \cellcolor{blue!25}1.75\tnote{1} & 16.49 & 5.75 & 25.8 & 1057 \\[1pt]
			\hline
			\hline
        	\noalign{\vskip 1pt}
		\end{tabular}
       	\begin{tablenotes}
     		\item[1] Blue denotes the best performance among all DL algorithms
		\end{tablenotes}
    \end{threeparttable}
\end{table}    

\subsection{Computational Efficiency 
} 
\label{subsec:comp_times}
We used MATLAB 2016a platform on an 8-Core CPU Windows 7 desktop PC to clock the times for DL algorithms. The ODL algorithm from \cite{mairal2009online} is implemented as \texttt{mex} executable, and therefore already fine-tuned for speed. For K-SVD, we employed the efficient implementation from \cite{rubinstein2008efficient} 
to improve computational speed. Table~\ref{tbl:times} lists the execution times of the five DL approaches. Here, the parameters were optimally selected for all the algorithms. The LRSDL is the slowest of all while ODL is more than 4 times faster than K-SVD. The CBWLSU provided better classification results but is three times slower than ODL. This could be because the dictionary update step always considers the entire previous training set elements that correlate with only one new element (i.e. there is no mini-batch strategy). This makes the convergence in CBWLSU more challenging.

The DOMINODL is the fastest DL method clocking 3x speed than ODL and 15x than K-SVD. This is because the DOMINODL updates the dictionary by evaluating only a mini-batch of previous elements (instead of all of them as in CBWLSU) that correlate with a mini-batch of several new elements (CBWLSU uses just one new element). Further, DOMINODL drops out the unused elements leading to a faster convergence. We note that, unlike ODL and K-SVD implementations, we did not use \texttt{mex} executables of DOMINODL which can further shorten current execution times. From Table~\ref{tbl:times}, the reduction in DOMINODL computational time over K-SVD is $((25.8-1.75)\times100)/25.8 \approx93$\%. The reduction for ODL and CBWLSU are computed similarly as $8$\% and $36$\%, respectively.

The computational bottleneck of mines classification lies in the training times. In comparison, the common steps of sparse decomposition and SVM-based classification during testing take just 0.4 s and 1 s, respectively, for an entire survey (1 m $\times$ 1 m area with 2500 range profiles). Thus, time taken per range profile in \texttildelow0.59 ms. The average scan rate of our GPR system is 0.19 m/s (or 1 cm/52.1 ms). This can go as high as 2.7 m/s (or 1 cm/3.61 ms) in other GPRs used for landmines application. Therefore, the test times do not impose much computational cost.

\subsection{Comparison with Sparse-Representation-Based Classification}
\begin{table}[ht]
\centering
	\begin{threeparttable}[b]
 		\caption{Confusion matrices using a coarse selection for $\mathbf{Y}$ including SRC}
		\label{tbl:CM_SRC_coarse}
		\begin{tabular}{ c | l | l | l | l | l}
			\hline
    		\noalign{\vskip 1pt}
            \multicolumn{2}{c|}{} & Clutter & PMN/PMA2 & ERA & Type-72\\[1pt]
        	\hline
        	\hline
        	\noalign{\vskip 1pt}
             						& Clutter & \cellcolor{black!25}0.912& 0.017 & 0.25 & 0.925 \\[1pt]
             \multirow{2}{*}{SRC} & PMN/PMA2 & 0.037 & \cellcolor{black!25}0.456\tnote{1} & 0.042 & 0 \\[1pt]
             						& ERA & 0.025 & 0.526 & \cellcolor{black!25}0.278 & 0.074 \\[1pt]
             						& Type-72 & 0.025 & 0 & 0.43 & \cellcolor{black!25}0 \\[1pt]
			\hline
        	\noalign{\vskip 1pt}
            					   	& Clutter & \cellcolor{black!25}0.729 & 0.017 & 0.083 & 0.241 \\[1pt]
            \multirow{2}{*}{DOMINODL} 	& PMN/PMA2 & 0.041 & \cellcolor{black!25}0.982 & 0.139 & 0 \\[1pt]
            					   	& ERA & 0.054 & 0 & \cellcolor{black!25}0.667 & 0.074 \\[1pt]
            					   	& Type-72 & 0.176 & 0 & 0.111 & \cellcolor{black!25}0.685 \\[1pt]
        	\hline
        	\noalign{\vskip 1pt}
           							& Clutter & \cellcolor{black!25}0.584 & 0 & 0.125 & 0.185 \\[1pt]
            \multirow{2}{*}{CBWLSU} & PMN/PMA2 & 0.063 & \cellcolor{black!25}0.982 & 0.153 & 0.11 \\[1pt]
            						& ERA & 0.106 & 0 & \cellcolor{black!25}0.625 & 0.185\\[1pt]
            						& Type-72 & 0.247 & 0.017 & 0.097& \cellcolor{black!25}0.518 \\[1pt]
			\hline
	    	\noalign{\vskip 1pt}
            						& Clutter & \cellcolor{black!25}0.71 & 0.035 & 0.153 & 0.259 \\[1pt]
            \multirow{2}{*}{ODL} & PMN/PMA2 & 0.036 & \cellcolor{black!25}0.912 & 0.069 & 0.074 \\[1pt]
            						& ERA & 0.088 & 0.008 & \cellcolor{black!25}0.667 & 0.074 \\[1pt]
            						& Type-72 & 0.165 & 0.044 & 0.111 & \cellcolor{black!25}0.593 \\[1pt]
			\hline
	    	\noalign{\vskip 1pt}
            						& Clutter & \cellcolor{black!25}0.617 & 0 & 0.111 & 0.148 \\[1pt]
            \multirow{2}{*}{K-SVD} & PMN/PMA2 & 0.044 & \cellcolor{black!25}0.982 & 0.194 & 0.056 \\[1pt]
            						& ERA & 0.113 & 0 & \cellcolor{black!25}0.667 & 0.241 \\[1pt]
            						& Type-72 & 0.226 & 0.017 & 0.027 & \cellcolor{black!25}0.444 \\[1pt]
 			\hline
			\hline
        	\noalign{\vskip 1pt}
		\end{tabular}
       	\begin{tablenotes}
			\item[1] Gray denotes the $\mathbf{P_{CC}}$ value for a specified class and DL algorithm
		\end{tablenotes}
    \end{threeparttable}
\end{table}    
We compared our proposed DL-based approach with the sparse-representation-based classification (SRC) method proposed in \cite{giovanneschi2015preliminary}. The SRC needs a labeled dictionary but the dictionary that we learn from $\mathbf{Y}$ does not have label information anymore thereby making SRC infeasible here. Therefore, we adopt the following steps for a reasonable comparison of the two methods. We feed SRC with $\mathbf{Y}$ as the dictionary $\mathbf{D}$. As indicated in Section IV.C, a meticulously selected collection of mines/clutter responses as $\mathbf{Y}$ is meaningful for comparing different DL approaches. But it does not highlight the benefits of employing DL \textit{per se}. Therefore, we generate a coarser selection, i.e. more profiles than the handpicked case, as $\mathbf{Y}$ for both approaches. Table~\ref{tbl:CM_SRC_coarse} shows the confusion matrix for the residual-based classification along with the proposed DL-based approaches. The DL-based mine classification is consistent with previous results - even better with DOMINODL - but with some trade-off of decreasing clutter accuracy. The accuracy of residual-based classifier is severely degraded for all mine classes, dropping by at least 45\%, 39\% and 55\% for PMA/PMA2, ERA and T72, respectively. This renders the increase in clutter classification accuracy of this method not usable.

\subsection{Deep-Learning-Based Classification}
\label{subsec:cnn}
\begin{figure}[t]
\centering
  \includegraphics[scale=0.35]{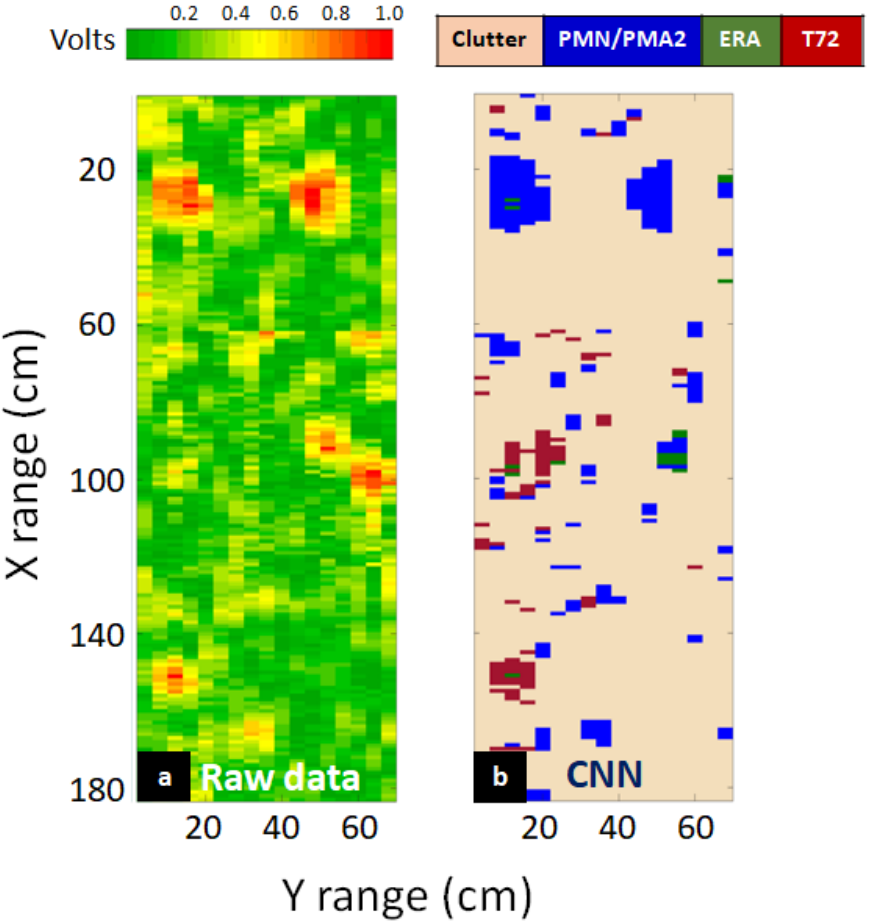}
  \caption{(a) Raw data at $15$ cm depth. (b) Classification maps of the same area containing 6 buried landmines using CNN-based classification.}
\label{fig:classmap_CNN}
\end{figure}
\begin{table}[t]
\centering
    \caption{Confusion matrix for CNN-based classification}
	\label{tbl:CNN_CM}
	\begin{tabular}{ l | l | l | l | l}
		\hline
         \noalign{\vskip 1pt}
            & Clutter & PMN/PMA2 & ERA & Type-72\\[1pt]
		\hline
        \hline
        \noalign{\vskip 1pt}
            Clutter & \cellcolor{black!25}0.909& 0.14 & 0.016 & 0.574 \\[1pt]
            PMN/PMA2 & 0.032 & \cellcolor{black!25}0.807 & 0.181 & 0 \\[1pt]
            ERA & 0 & 0.053 & \cellcolor{black!25}0.319 & 0.315 \\[1pt]
            Type-72 & 0.033 & 0 & 0.111 & \cellcolor{black!25}0.370 \\[1pt]
		\hline
		\hline
        \noalign{\vskip 1pt}
	\end{tabular}
\end{table}   
The core idea of SR-based classification is largely based on the assumption that signals are linear combinations of a few atoms. In practice, this is often not the case. This has led to a few recent works \cite{vu2018deep} that suggest employing deep learning for radar target classification. However, these techniques require significantly large datasets for training.

We compared classification results of our methods with a deep learning approach. In particular, we constructed a CNN because these networks are known to efficiently exploit structural or locational information in the data and yield comparable learning potential with far fewer parameters \cite{girshick2015fast}. We modeled our proposed CNN framework as a classification problem wherein each class denotes the type of mine or clutter. The training data set for our CNN structure is the matrix $\mathbf{Y}$ (see Section~\ref{sec:meas_camp}). Building up a synthetic database is usually an option for creating (or extending) a training set for deep learning applications. However, accurately modeling a GPR scenario is still an ongoing challenge in the GPR community because of the difficulties in accurately reproducing the soil inhomogeneities (and variabilities), the surface and underground clutter, the antenna coupling and ringing effects, etc. Even though some applications have been promising \cite{giannakis2016realistic}, this remains a cumbersome task. 

The input layer of our CNN took one-dimensional sample set of size $211$. It was followed by two convolutional layers with $20$ and $5$ filters of size $20$ and $10$, respectively. The output layer consisted of four units wherein the network classifies the given input data as clutter or one of the three mines. There were rectified linear units (ReLU) after each convolutional layer; the ReLU function is given by $\text{ReLU}(x) = \text{max}(x,0)$ \cite{srivastavaDropoutLayer}.
The architecture of the CNN was selected through an arduous process of testing many combination of layers/filters and hyperparameters which would lead to better accuracy during training. A deeper network slightly increased the accuracy in the training phase but led to poorer performance when classifying new data (i.e. the test set $\mathbf{Y}_{test}$). Since our data are limited, adding more layers (i.e. more weights) only led to overfitting and made the network incapable to generalize on new datasets. A multi-dimensional CNN formed by clustering 2D and 3D data would have further reduced the training set. Augmenting the data was also envisioned but commonly used transformations such as scaling/rotations are not useful in our case because the mines were always in the same inclination and their dimension defines the class itself. We also attempted adding different levels of noise but this did not lead to better results considering the available data are already very noisy.
\begin{figure*}
\centering
  \includegraphics[scale=0.50]{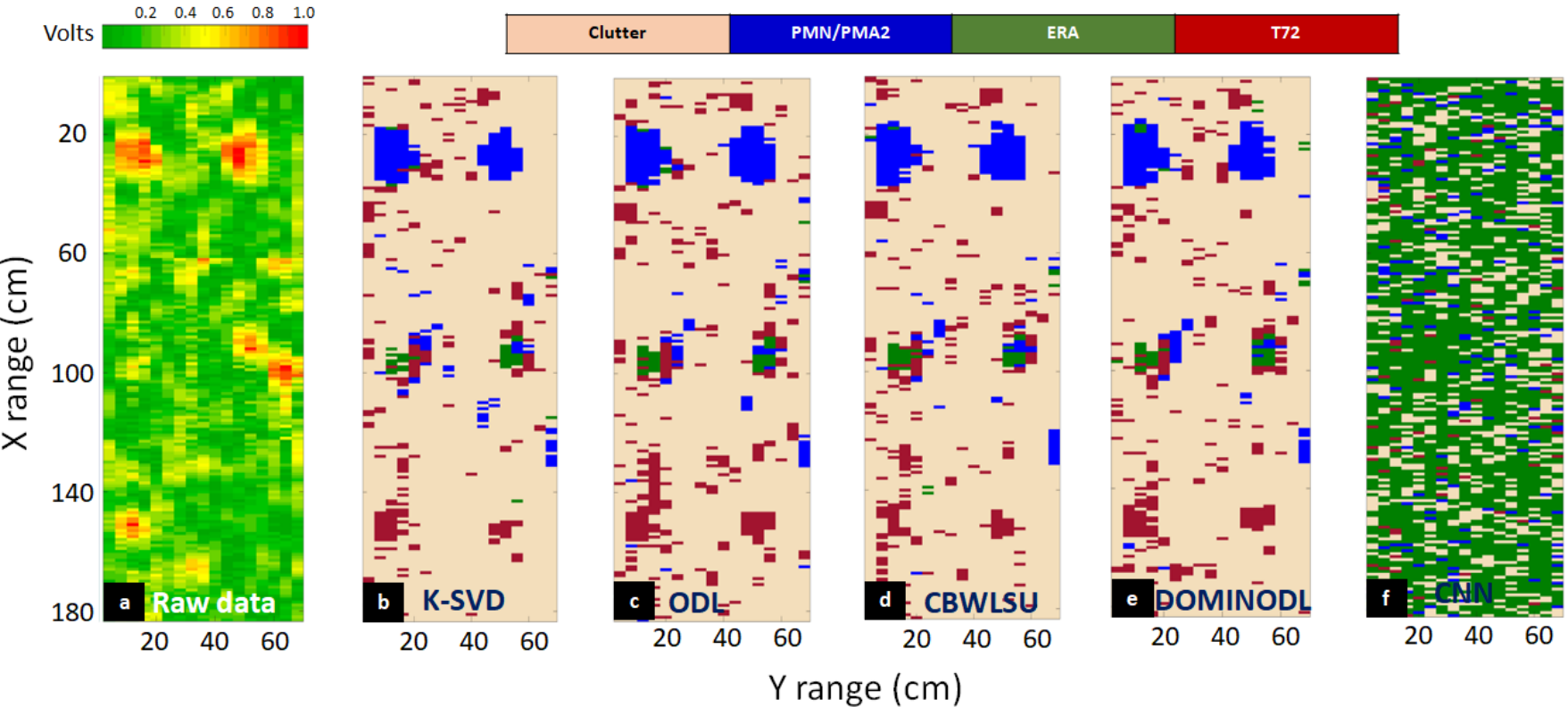}
  \caption{(a) Raw data at $15$ cm depth. The classification maps of the same area containing 6 buried landmines using an SR-based approach with dictionary learned using (b) K-SVD, (c) ODL, (d) CBWLSU, and (e) DOMINODL algorithms. The input parameters were optimally selected and the number of samples were reduced by $50$\%. (f) The corresponding result with reduced samples for CNN-based classification.}
\label{fig:classmap_50}
\end{figure*}
\begin{table*}
\centering
    \caption{Confusion matrices for different DL algorithms and CNN with reduced signal samples}
	\label{tbl:cm_sensing_less}
		\begin{tabular}{ p{1.2cm}| p{1.1cm} ? l | p{1.1cm} | l | l ?? l | p{1.1cm} | l | l ?? l | p{1.1cm} | l | l }
		\hline
    	\noalign{\vskip 1pt}
            \multicolumn{2}{c?}{\multirow{2}{*}{}} & \multicolumn{4}{c??}{25\% Reduction} & \multicolumn{4}{c??}{50\% Reduction} & \multicolumn{4}{c}{75\% Reduction}\\[1pt]
             \cline{3-14}
        \noalign{\vskip 1pt}
            \multicolumn{2}{c?}{} & Clutter & PMN/PMA2 & ERA & Type-72 & Clutter & PMN/PMA2 & ERA & Type-72 & Clutter & PMN/PMA2 & ERA & Type-72\\[1pt]
        \hline
        \hline
        \noalign{\vskip 1pt}
            						& Clutter & \cellcolor{black!25}0.892& 0.078 & 0.319 & 0.389 & \cellcolor{black!25}0.882& 0.026 & 0.291 & 0.37 & \cellcolor{black!25}0.877 & 0.061 & 0.402 & 0.426\\[1pt]
            \multirow{2}{*}{K-SVD} 	& PMN/PMA2 & 0.021 & \cellcolor{black!25}0.921 & 0.153 & 0.055 & 0.018 & \cellcolor{black!25}0.947 & 0.153 & 0.037 & 0.02 & \cellcolor{black!25}0.912 & 0.125 & 0.074\\[1pt]
            						& ERA & 0.021 & 0 & \cellcolor{black!25}0.486 & 0.018 & 0.021 & 0.026 & \cellcolor{black!25}0.5 & 0 & 0.021 & 0.026 & \cellcolor{black!25}0.333 & 0.018\\[1pt]
            						& Type-72 & 0.065 & 0 & 0.041 & \cellcolor{black!25}0.537 & 0.078 & 0 & 0.055 & \cellcolor{black!25}0.592 & 0.08 & 0 & 0.138 & \cellcolor{black!25}0.481\\[1pt]
		\hline
        \noalign{\vskip 1pt}
            						& Clutter & \cellcolor{black!25}0.872 & 0.088 & 0.208 & 0.315 & \cellcolor{black!25}0.868 & 0 & 0.208 & 0.333 & \cellcolor{black!25}0.862 & 0.02 & 0.319 & 0.296\\[1pt]
            \multirow{2}{*}{ODL} 	& PMN/PMA2 & 0.021 & \cellcolor{black!25}0.973 & 0.152 & 0 & 0.021 & \cellcolor{black!25}0.965 & 0.18 & 0.018 & 0.023 & \cellcolor{black!25}0.964 & 0.138 & 0.018\\[1pt]
            						& ERA & 0.018 & 0.017 & \cellcolor{black!25}0.527 & 0.018 & 0.018 & 0.035 & \cellcolor{black!25}0.5 & 0 & 0.021 & 0.008 & \cellcolor{black!25}0.416 & 0.074\\[1pt]
            						& Type-72 & 0.087 & 0 & 0.111 & \cellcolor{black!25}0.666 & 0.09 & 0 & 0.111 & \cellcolor{black!25}0.648 & 0.091 & 0 & 0.125 & \cellcolor{black!25}0.611\\[1pt]
        \hline
		\noalign{\vskip 1pt}
            						& Clutter & \cellcolor{black!25}0.871 & 0.026 & 0.194 & 0.351 & \cellcolor{black!25}0.872 & 0.017 & 0.25 & 0.40 & \cellcolor{black!25}0.855 & 0.088 & 0.388 & 0.370\\[1pt]
            \multirow{2}{*}{CBWLSU} & PMN/PMA2 & 0.024 & \cellcolor{black!25}0.956 & 0.139 & 0 & 0.023 & \cellcolor{black!25}0.973 & 0.111 & 0 & 0.024 & \cellcolor{black!25}0.974 & 0.111 & 0\\[1pt]
            						& ERA & 0.025 & 0.017 & \cellcolor{black!25}0.541 & 0 & 0.02 & 0.008 & \cellcolor{black!25}0.541 & 0 & 0.027 & 0.017 & \cellcolor{black!25}0.333 & 0.018\\[1pt]
            						& Type-72 & 0.79 & 0 & 0.125 & \cellcolor{black!25}0.648 & 0.083 & 0 & 0.097 & \cellcolor{black!25}0.592 & 0.091 & 0 & 0.125 & \cellcolor{black!25}0.611\\[1pt]
        \hline
        \noalign{\vskip 1pt}
            						& Clutter & \cellcolor{black!25}0.88 & 0.017 & 0.236 & 0.277 & \cellcolor{black!25}0.868 & 0.035 & 0.194 & 0.296 & \cellcolor{black!25}0.864 & 0.035 & 0.278 & 0.444\\[1pt]
            \multirow{2}{*}{DOMINODL} & PMN/PMA2 & 0.022 & \cellcolor{black!25}0.964 & 0.138 & 0 & 0.023 & \cellcolor{black!25}0.929 & 0.138 & 0 & 0.027 & \cellcolor{black!25}0.938 & 0.152 & 0\\[1pt]
            						& ERA & 0.018 & 0.017 & \cellcolor{black!25}0.527 & 0 & 0.024 & 0.035 & \cellcolor{black!25}0.527 & 0.018 & 0.026 & 0.026 & \cellcolor{black!25}0.5 & 0\\[1pt]
            						& Type-72 & 0.078 & 0 & 0.097 & \cellcolor{black!25}0.722 & 0.083 & 0 & 0.138 & \cellcolor{black!25}0.685 & 0.082 & 0 & 0.069 & \cellcolor{black!25}0.556\\[1pt]
        \hline
        \noalign{\vskip 1pt}
            						& Clutter & \cellcolor{black!25}0.708 & 0.359 & 0.236 & 0.407 & \cellcolor{black!25}0.265 & 0.166 & 0.645 & 0.148 & \cellcolor{black!25}0.162 & 0.105 & 0.647 & 0.129\\[1pt]
            \multirow{2}{*}{CNN} 	& PMN/PMA2 & 0.026 & \cellcolor{black!25}0.41 & 0.097 & 0.018 & 0.062 & \cellcolor{black!25}0.096 & 0.069 & 0.018 & 0.015 & \cellcolor{black!25}0.061 & 0.013 & 0\\[1pt]
            						& ERA & 0 & 0.21 & \cellcolor{black!25}0.5 & 0.426 & 0 & 0.72 & \cellcolor{black!25}0.73 & 0.75 & 0 & 0.71 & \cellcolor{black!25}0.708 & 0.759\\[1pt]
            						& Type-72 & 0.029 & 0.017 & 0.069 & \cellcolor{black!25}0.148 & 0.027 & 0.088 & 0.014 & \cellcolor{black!25}0.074 & 0.17 & 0.12 & 0.12 & \cellcolor{black!25}0.11\\[1pt]
        \hline
		\hline
        \noalign{\vskip 1pt}
	\end{tabular}
\end{table*}    
\begin{table*}
\centering
    \caption{Confusion matrices for different DL algorithms and CNN with reduced training set elements}
	\label{tbl:cm_sensing_less_te2}
		\begin{tabular}{ p{1.2cm}| p{1.1cm} ? l | p{1.1cm} | l | l ?? l | p{1.1cm} | l | l ?? l | p{1.1cm} | l | l }
		\hline
    	\noalign{\vskip 1pt}
            \multicolumn{2}{c?}{\multirow{2}{*}{}} & \multicolumn{4}{c??}{25\% Reduction} & \multicolumn{4}{c??}{50\% Reduction} & \multicolumn{4}{c}{75\% Reduction}\\[1pt]
             \cline{3-14}
        \noalign{\vskip 1pt}
            \multicolumn{2}{c?}{} & Clutter & PMN/PMA2 & ERA & Type-72 & Clutter & PMN/PMA2 & ERA & Type-72 & Clutter & PMN/PMA2 & ERA & Type-72\\[1pt]
        \hline
        \hline
        \noalign{\vskip 1pt}
            						& Clutter & \cellcolor{black!25}0.887& 0.044 & 0.236 & 0.370 & \cellcolor{black!25}0.868& 0 & 0.417 & 0.556 & \cellcolor{black!25}0.885 & 0.097 & 0.444 & 0.426\\[1pt]
            \multirow{2}{*}{K-SVD} 	& PMN/PMA2 & 0.019 & \cellcolor{black!25}0.912 & 0.139 & 0 & 0.023 & \cellcolor{black!25}0.912 & 0.153 & 0.037 & 0.019 & \cellcolor{black!25}0.746 & 0.014 & 0.130\\[1pt]
            						& ERA & 0.023 & 0.044 & \cellcolor{black!25}0.528 & 0.019 & 0.010 & 0.070 & \cellcolor{black!25}0.306 & 0.019 & 0.012 & 0.088 & \cellcolor{black!25}0.319 & 0 \\[1pt]
            						& Type-72 & 0.071 & 0 & 0.097 & \cellcolor{black!25}0.611 & 0.099 & 0.018 & 0.125 & \cellcolor{black!25}0.389 & 0.084 & 0.070 & 0.222 & \cellcolor{black!25}0.444\\[1pt]
		\hline
        \noalign{\vskip 1pt}
            						& Clutter & \cellcolor{black!25}0.865 & 0.009 & 0.167 & 0.315 & \cellcolor{black!25}0.865 & 0.009 & 0.375 & 0.611 & \cellcolor{black!25}0.857 & 0.070 & 0.375 & 0.352\\[1pt]
            \multirow{2}{*}{ODL} 	& PMN/PMA2 & 0.016 & \cellcolor{black!25}0.939 & 0.181 & 0 & 0.023 & \cellcolor{black!25}0.991 & 0.111 & 0 & 0.027 & \cellcolor{black!25}0.851 & 0.056 & 0.093\\[1pt]
            						& ERA & 0.020 & 0.053 & \cellcolor{black!25}0.583 & 0.019 & 0.011 & 0 & \cellcolor{black!25}0.431 & 0 & 0.016 & 0.079 & \cellcolor{black!25}0.444 & 0\\[1pt]
            						& Type-72 & 0.096 & 0 & 0.069 & \cellcolor{black!25}0.667 & 0.101 & 0 & 0.083 & \cellcolor{black!25}0.389 & 0.101 & 0 & 0.125 & \cellcolor{black!25}0.556\\[1pt]
        \hline
		\noalign{\vskip 1pt}
            						& Clutter & \cellcolor{black!25}0.865 & 0.035 & 0.250 & 0.426 & \cellcolor{black!25}0.874 & 0.009 & 0.347 & 0.407 & \cellcolor{black!25}0.834 & 0.035 & 0.500 & 0.685\\[1pt]
            \multirow{2}{*}{CBWLSU} & PMN/PMA2 & 0.021 & \cellcolor{black!25}0.912 & 0.167 & 0 & 0.033 & \cellcolor{black!25}0.904 & 0.069 & 0.241 & 0.020 & \cellcolor{black!25}0.851 & 0.056 & 0 \\[1pt]
            						& ERA & 0.026 & 0.053 & \cellcolor{black!25}0.542 & 0.019 & 0.014 & 0.088 & \cellcolor{black!25}0.500 & 0 & 0.015 & 0.105 & \cellcolor{black!25}0.403 & 0\\[1pt]
            						& Type-72 & 0.088 & 0 & 0.042 & \cellcolor{black!25}0.556 & 0.080 & 0 & 0.083 & \cellcolor{black!25}0.352 & 0.131 & 0.089 & 0.042 & \cellcolor{black!25}0.315\\[1pt]
        \hline
        \noalign{\vskip 1pt}
            						& Clutter & \cellcolor{black!25}0.841 & 0.026 & 0.222 & 0.296 & \cellcolor{black!25}0.864 & 0.053 & 0.333 & 0.574 & \cellcolor{black!25}0.815 & 0.175 & 0.431 & 0.519\\[1pt]
            \multirow{2}{*}{DOMINODL} & PMN/PMA2 & 0.021 & \cellcolor{black!25}0.921 & 0.167 & 0 & 0.023 & \cellcolor{black!25}0.868 & 0.069 & 0.074 & 0.027 & \cellcolor{black!25}0.746 & 0.069 & 0\\[1pt]
            						& ERA & 0.026 & 0.053 & \cellcolor{black!25}0.542 & 0.019 & 0.014 & 0.079 & \cellcolor{black!25}0.486 & 0 & 0.017 & 0.079 & \cellcolor{black!25}0.486 & 0\\[1pt]
            						& Type-72 & 0.112 & 0 & 0.070 & \cellcolor{black!25}0.685 & 0.099 & 0 & 0.111 & \cellcolor{black!25}0.352 & 0.142 & 0 & 0.014 & \cellcolor{black!25}0.482\\[1pt]
        \hline
        \noalign{\vskip 1pt}
            						& Clutter & \cellcolor{black!25}0.838 & 0.026 & 0.040 & 0.333 & \cellcolor{black!25}0.819 & 0.088 & 0.047 & 0.315 & \cellcolor{black!25}0.644 & 0.035 & 0.109 & 0.463 \\[1pt]
            \multirow{2}{*}{CNN} 	& PMN/PMA2 & 0.062 & \cellcolor{black!25}0.868 & 0.208 & 0.537 & 0.059 & \cellcolor{black!25}0.851 & 0.319 & 0.463 & 0.080 & \cellcolor{black!25}0.693 & 0.181 & 0.019 \\[1pt]
            						& ERA & 0 & 0.053 & \cellcolor{black!25}0.250 & 0.074 & 0 & 0 & \cellcolor{black!25}0.153 & 0.130 & 0 & 0.193 & \cellcolor{black!25}0.306 & 0.019 \\[1pt]
            						& Type-72 & 0.106 & 0.053 & 0.208 & \cellcolor{black!25}0.556 & 0.120 & 0.061 & 0.250 & \cellcolor{black!25}0.500 & 0.105 & 0.079 & 0.097 & \cellcolor{black!25}0\\[1pt]
        \hline
		\hline
        \noalign{\vskip 1pt}
	\end{tabular}
\end{table*}    
We trained the network with the labeled training set $\mathbf{Y}$, selecting \texttildelow$20$\% of the training data for validation. Specifically, the validation set employed $100$, $25$, $25$, and $25$ range profiles for clutter, PMN/PMA2, ERA and Type-72, respectively. We used a stochastic gradient descent algorithm for updating the network parameters with the learning rate of $0.001$ and mini-batch size of $20$ samples for $2000$ epochs.

We realized the proposed network in TensorFlow on a Windows 7 PC with 8-core CPU. The network training took $3.88$ minutes. Figure~\ref{fig:classmap_CNN} shows the classification map obtained using CNN. The corresponding confusion matrix is listed in Table~\ref{tbl:CNN_CM}. We note that the CNN classifier shows worse $P_{CC}$ than our SR-based techniques, particularly for ERA and Type-72 target classes.  

\subsection{Classification with Reduced Range Samples}
\label{subsec:sparse_sensing}
We now analyze the robustness of our DL-based adaptive classification method to the reduction of the number of samples in the raw data. Assuming the collected data $\mathbf{Y}_{\text{TEST}}$ is sparse in dictionary $\mathbf{D}$, we undersampled the original raw data $\mathbf{Y}_{\text{TEST}}$ in range to obtain its row-undersampled version $\widetilde{\mathbf{Y}}_{\text{TEST}}$ by randomly reducing the samples. We then applied the same random sampling pattern to the dictionary $\mathbf{D}$ for obtaining the sparse coefficients. We also analyzed the CNN classifier when the signals are randomly reduced in the same way. Figure~\ref{fig:classmap_50} illustrates the classification map for all DL approaches when the sampling is reduced by $50$\%. Table~\ref{tbl:cm_sensing_less} clubs together the confusion matrices when undersampling by $25$\%, $50$\%, and $75$\%.

In comparison to the results in Table~\ref{tbl:CM_DL_full} which used all samples of the raw data, the DL approaches maintain similar classifier performance even when we reduce the samples by 75\% (i.e. just 52 samples in total). In contrast, the CNN classifier result which is already heavily compromised with a reduction of $25$\%, fails completely for $50$\%and $75$\% sampling rate. Reducing the number of signal samples when using a dictionary which minimizes the number of non-zero entries in the sparse representation, still assures an exact reconstruction of the signal itself and, consequently its correct classification. The features for classifying the traces are robust to the reduction of the original samples. Deep learning strategies use the signal samples directly as classification features. They also require enormous amount of data for training. Therefore, the degradation in their performance is expected. From the confusion matrix in Table~\ref{tbl:cm_sensing_less} indicates that CNN has the highest $P_{CC}$ for ERA. This is a false trail because the network mis-classified almost every pixel as ERA. Overall, DOMINODL and CBWLSU provide excellent results for small mines. However, as seen earlier, CBWLSU is not very well-suited for real-time operation because of longer execution times.

We also assessed the performance of different methods when, instead of sampling fewer range samples per profile, we include all samples in every range profile but reduce the overall number of training set elements randomly from 926 to 694, 464 and 232 range profiles (which respectively correspond to 25\%, 50\% and 75\% reduction). From the corresponding confusion matrices listed in Table~\ref{tbl:cm_sensing_less_te2}, we note that CNN-based classification results have improved with respect to Table~\ref{tbl:cm_sensing_less}. However, the classification accuracy of CNN is still poorer than the DL-based classification. In general, all methods show performance degradation as the training set elements are reduced. Among the online DL methods, ODL is more robust to the range profile reduction than K-SVD.

\section{Summary}
\label{sec:summary}
In this paper, we proposed effective online DL strategies for sparse decomposition of GPR traces of buried landmines. The online methods outperform K-SVD thereby making them a good candidate for SR-based classification. Our algorithm DOMINODL is always the fastest providing near real-time performance and high clutter rejection while also maintaining a classifier performance that is comparable to other online DL algorithms. DOMINODL and CBWLSU generally classify smaller mines better than ODL and K-SVD. Unlike previous works that rely on RMSE, we used metrics based on statistical inference to tune the DL parameters for enhanced operation.

Fast ODL computations pave the way towards cognition \cite{mishra2018sub,mishra2016cognitive,mishra2017performance} in GPR operation, wherein the system uses previous measurements to optimize the processing performance and is capable of sequential sampling adaptation \cite{mishra2014compressed} based on the learned dictionary. For example, in a realistic landmine clearance campaign, an operator could gather the training measurements over a safe area next to the contaminated site, hypothetically placing some buried landmine simulants over it in order to have a faithful representation of the soil/targets interaction beneath the surface. In other words, our work allows the operator to \textit{calibrate} the acquisition by providing a good training set to learn the dictionary.

\section*{Acknowledgements}
\label{sec:ack}
The authors acknowledge valuable assistance from David Mateos-N\'{u}\~{n}ez in Section~\ref{subsec:cnn}.

\bibliographystyle{IEEEtran}
\bibliography{refs}

\end{document}